\documentclass{article}

\usepackage{amsmath}
\usepackage{amssymb}
\usepackage{algorithm}
\usepackage{algpseudocode}
\usepackage{wrapfig}
\usepackage{needspace}
\usepackage[numbers]{natbib}
\usepackage[main, final]{neurips_2026}
\usepackage{graphicx}
\usepackage{bm}

\usepackage{booktabs}
\usepackage{multirow}
\usepackage[table,dvipsnames]{xcolor}
\usepackage{array}
\usepackage{graphicx}

\definecolor{quanviblue}{RGB}{235,245,255}

\usepackage[utf8]{inputenc} 
\usepackage[T1]{fontenc}    
\usepackage{hyperref}       
\usepackage{url}            
\usepackage{booktabs}       
\usepackage{amsfonts}       
\usepackage{nicefrac}       
\usepackage{microtype}      
\usepackage{subcaption}

\usepackage{multirow}

\newcommand{\model}{QuanVI}

\hypersetup{
colorlinks=true,
citecolor=Blue,
linkcolor=Green,
urlcolor=Red,
}

\title{Quantum-Inspired Score-Based Variational Inference}

\author{%
  Yuchen Cong\\
  Juntendo University
  \And
  Zerui Tao\\
  RIKEN-AIP
  \And
  Chao Li%
  \thanks{\texttt{chao.li@riken.jp}}\\
  RIKEN-AIP
  \AND
  Zhe Sun\\
  Juntendo University
  \And
  Qibin Zhao\\
  RIKEN-AIP
}

\begin{document}

\maketitle

\begin{abstract}
Score-based variational inference (SBVI) provides an alternative
to classical Kullback--Leibler (KL)-based variational inference
by learning a distribution using only score queries of the target.
EigenVI~\cite{cai2024eigenvi} formulates SBVI as an eigenvalue
problem, but faces two limitations:
the number of variational parameters grows exponentially with the data dimension, and the single-eigenvector formulation cannot represent the full minimum-eigenvalue eigenspace.
We propose \emph{\model{}}, a quantum-inspired SBVI algorithm
that addresses both limitations.
Rather than selecting a single eigenvector,
\model{} captures a learned subspace through a density-operator formulation.
To enable scalable computation, \model{} employs a
matrix-product-operator (MPO) structure parameterized
by a quantum tensor network (QTN) with unitary-constrained
local tensors and auxiliary ancilla wires.
Numerical experiments demonstrate the scalability of
\model{} beyond EigenVI and competitive performance on
synthetic distributions and Bayesian posterior-approximation
benchmarks, including challenging non-Gaussian targets.
The code is available at \url{https://github.com/ChaoLiAtRIKEN/quanvi}.

\end{abstract}

\section{Introduction}
\label{sec:intro}
Variational inference (VI, \citep{blei2017variational,wainwright2008graphical}) approximates a target distribution by finding the closest member of a parameterized family of distributions. 
Conventional VI methods minimize the Kullback--Leibler (KL) divergence \cite{kullback1951information} between the variational approximation and the target distribution.
In contrast, score-based VI (SBVI) minimizes the Fisher divergence between the two distributions by matching their score functions. 
Consequently, SBVI is particularly attractive when target score evaluations are available but target density evaluations are computationally expensive or intractable.

Previous studies have investigated SBVI using Gaussian variational families~\citep{modi2023variational,caibatch}.
Recently, EigenVI~\citep{cai2024eigenvi} introduced non-Gaussian variational families based on orthogonal basis expansions.
More intriguingly, the work reveals that minimizing Fisher divergence with EigenVI reduces to solving a minimum-eigenvalue problem.
Despite its promising performance, EigenVI faces two limitations:
first, the number of variational parameters grows \emph{exponentially} with the data dimension, limiting the scalability of EigenVI;
second, the single-eigenvector formulation does not account for the possible \emph{multiplicity} of the minimum eigenvalue, potentially limiting the expressive power of the variational family.



To address these limitations, we propose \emph{\model{}}, a novel SBVI method inspired by quantum computing.
Building on EigenVI, \model{} extends the variational parameters from a single vector to a density operator, allowing the model to capture multiple eigen-directions associated with the minimum eigenvalue.
To avoid the exponential growth of parameters, we further represent the density operator as a matrix product operator (MPO)~\citep{verstraete2004matrix} using quantum tensor networks (QTNs).
The QTN representation decomposes the high-dimensional density operator into multiplications of low-dimensional local tensors, analogous to quantum circuits.
With additional unitary constraints on the local tensors, \model{} enables tractable probabilistic inference without explicitly computing the partition function during training or inference.

In addition,  \model{} addresses the high query complexity of estimating the loss-induced Hamiltonian $M$ in~\eqref{eq:cai_eigen} due to the importance sampling, as required by EigenVI.
Under a bounded local dependency assumption, such as a Markov random field with bounded neighborhood size, the Hamiltonian can be represented as a sum of local terms, enabling importance sampling in low-dimensional spaces with reduced query complexity.

Empirically, we evaluate \model{} on synthetic distributions to examine its scalability and expressiveness, and on Bayesian posterior-approximation benchmarks from \textsc{PosteriorDB}~\citep{magnusson2024posteriordb} to assess its practical applicability.
We also conduct ablation studies to examine the effects of various hyperparameters.
The results demonstrate that \model{} scales to dimensions beyond the reach of EigenVI while achieving competitive approximation quality on Gaussian, non-Gaussian, and Bayesian posterior-approximation tasks.



\section{Preliminaries}
\label{sec:Pre}
In this section, we review EigenVI and the graphical notation of QTNs.

\subsection{EigenVI and Its Eigenvalue Formulation}
\label{eigenvi}

In EigenVI, a variational density $q_\theta:\mathbb{R}^D\rightarrow{}\mathbb{R}_{\geq{}0}$ is constructed by an orthogonal function expansion as follows:
\begin{equation}
    q_\theta(x)
    =
    \left|
    \left\langle
        \theta,
        \Phi(x)
    \right\rangle
    \right|^2,
    \qquad
    \Phi(x)=\bigotimes_{d=1}^D \phi(x_d),
    \label{eq:cai_model}
\end{equation}
where $\theta\in\mathbb{R}^{K^D}$ denotes a vector corresponding to the variational parameters, $\otimes$ denotes the tensor product, and $\phi:\mathbb{R}\rightarrow\mathbb{R}^K$ represents a vector of basis functions, satisfying the orthonormal assumption $\int \phi(t)\phi(t)^\top\,dt = I_K$.
If we further assume $\Vert{}\theta\Vert_F=1$, it can be easily derived that $q_\theta$ should be a valid density function, since the square operation in~\eqref{eq:cai_model} guarantees the non-negativity of the range of $q_\theta(x)$, and normalized $\theta$ with orthogonal basis functions makes the integral of $q_\theta(x)$ wrt. $x$ equals one.
In the TN community, ~\eqref{eq:cai_model} is also often called continuous-variable Born machine~\citep{meiburg2025generative}.

Suppose that a target density $p$ is unknown but accessible by its score function $\nabla_x\log{}p(x)$.
EigenVI aims to solve the following optimization problem:
\begin{equation}
    \min_\theta
    \frac{1}{2}
    \int q_\theta(x)
    \left\|
        \nabla_x\log p(x)-\nabla_x\log q_\theta(x)
    \right\|_F^2 dx,\qquad{}{}\mbox{s.t.}\,\Vert{}\theta\Vert_F=1.
    \label{eq:fisher}
\end{equation}
We can see from~\eqref{eq:fisher} that EigenVI minimizes the Fisher divergence~\citep{hyvarinen2005estimation} between $p$ and $q_\theta$.

An important observation established in EigenVI is that solving~\eqref{eq:fisher} is equivalent to solving a minimum-eigenvalue problem:
\begin{equation}
    \min_{\theta} \theta^\top M\theta,
    \qquad
    \text{s.t. } \|\theta\|_F^2=1,
    \label{eq:cai_eigen}
\end{equation}
where $M\in\mathbb{R}^{K^D\times K^D}$ denotes a loss-induced matrix defined as follows:
\begin{equation}
    M
    :=
    \int
    \left(
        2\dot{\Phi}(x)
        -
        \nabla_x\log{}p(x)\cdot\Phi(x)^\top
    \right)^\top
    \left(
        2\dot{\Phi}(x)
        -
        \nabla_x\log{}p(x)\cdot\Phi(x)^\top
    \right)
    dx,
    \label{eq:def_M}
 \end{equation}
where $\dot{\Phi}(x)$ denotes a Jacobian-like form of $\Phi(x)$, explicitly defined in~\eqref{eq:app_dot_phi}.
The derivation of $M$ from~\eqref{eq:fisher} is provided in Appendix~\ref{app:eigenvi_details}.
Hence, an optimal $\theta$ should belong to the eigenvector wrt. the minimum eigenvalue of $M$.
Any eigensolver can be applied to tackle the task.

\paragraph{Limitations of EigenVI.}
Directly solving~\eqref{eq:cai_eigen} faces three main challenges.
First, EigenVI selects a single eigenvector associated with the minimum eigenvalue of $M$, overlooking the possibility that this eigenvalue has multiplicity greater than one.
This restriction may limit the expressiveness of the variational family.
Second, the dimension of $\theta$ grows exponentially with the data dimension $D$, making the eigenvalue problem in~\eqref{eq:cai_eigen} computationally intractable in high dimensions.
Third, EigenVI estimates $M$ using importance sampling over the global $D$-dimensional space, which can require exponentially many target score queries as $D$ increases.
These limitations hinder the practical application of EigenVI to high-dimensional problems, despite its promising performance on non-Gaussian distributions~\citep{cai2024eigenvi}.

\begin{figure}[t]
    \centering
    \begin{subfigure}[t]{0.46\linewidth}
        \centering
        \includegraphics[width=.85\linewidth]{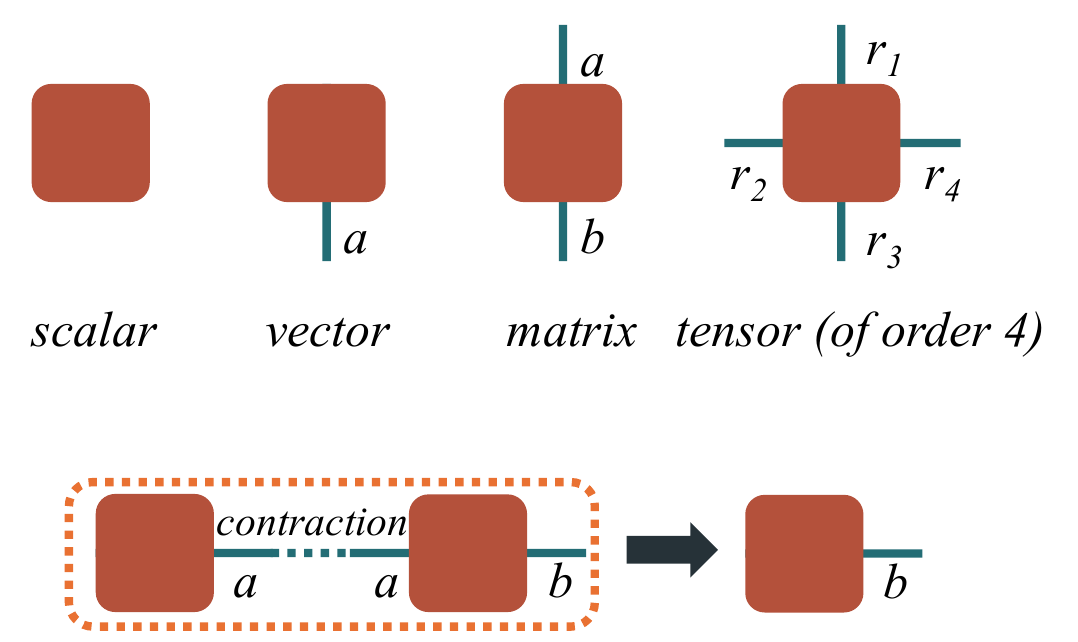}
        \caption{Basics of TN graphical notation. 
        Each node represents a tensor, and each edge represents an index (mode) of the tensor. 
        The label on an edge indicates the dimension of the corresponding mode. 
        }
        \label{fig:tn_contraction_example}
    \end{subfigure}
    \hspace{2em}
    \begin{subfigure}[t]{0.46\linewidth}
        \centering
        \includegraphics[width=.55\linewidth]{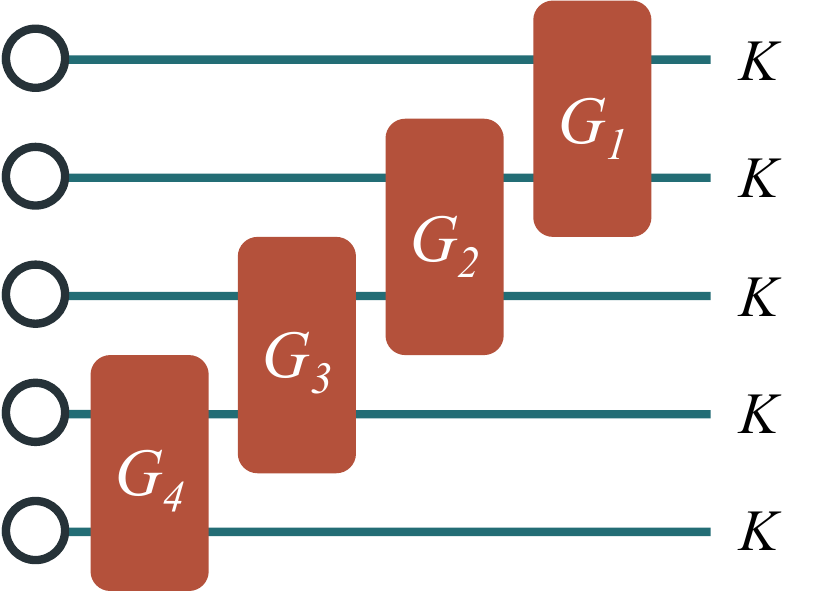}
        \caption{A simple QTN example.
        A QTN is a circuit-like TN architecture with additional unitary constraints on local tensors.
        In this example, circles represent initial-state tensors (vector) of dimension $K$, while rectangles denote local tensors of dimensions $K \times K \times K \times K$.
        }
        \label{fig:qtn_example}
    \end{subfigure}

    \caption{Graphical illustration of tensor-network notation and representations.
    (a) Basic tensor objects and tensor contraction.
    (b) A simple QTN example.}
    \label{fig:tn_example}
    \vspace{-0.8em}
\end{figure}

\subsection{Quantum Tensor Networks and Graphical Notation}
\label{sec:tn_qc}

To overcome the limitations of EigenVI, \model{} builds on tensor networks, which represent high-dimensional tensors through contractions of smaller local tensors.
Figure~\ref{fig:tn_contraction_example} illustrates the graphical notation for tensor contractions.
Each node represents a tensor, each edge represents an index, and connected edges indicate contraction.
The dimension of a shared index is called the bond dimension.
We use this notation throughout the paper to illustrate tensor operations in \model{}.

We focus on quantum tensor networks (QTNs), a class of tensor networks inspired by quantum circuits, as illustrated in Figure~\ref{fig:qtn_example}.
In our QTN architecture, local tensors consist of \emph{initial-state tensors} (shown as circles) and unitary \emph{gate tensors} (shown as rectangles).
Tensor contractions are represented by \emph{wires}, as in quantum circuits.
The unitary constraints preserve normalization, eliminating the need for an explicit partition function when combined with orthogonal basis functions.
Moreover, the structured contractions in our QTN architecture enable efficient evaluation of marginal probabilities.



\section{Method}
\label{method}

We now introduce \model{}, which builds on three key ideas.
First, we extend the variational vector $\theta$ in EigenVI to a density operator.
Second, we use a QTN to represent the density operator efficiently.
Third, we exploit local dependencies to express the high-dimensional loss-induced matrix $M$ as a sum of local terms.
The density-operator formulation allows multiple eigen-directions associated with the minimum eigenvalue, the QTN parameterization avoids exponentially many explicit variational parameters, and the local decomposition reduces the query complexity of estimating $M$.

\subsection{Extending EigenVI with Density Operators}
\label{sec:density_operator}

Recall the EigenVI optimization problem in~\eqref{eq:cai_eigen}.
We generalize this problem to
\begin{equation}
\begin{aligned}
\min_{\rho}\quad & \operatorname{tr}(\rho M),\
\text{s.t.}\quad
& \rho\succeq 0,\quad
\operatorname{tr}(\rho)=1,\quad
\operatorname{rank}(\rho)\leq r,
\end{aligned}
\label{eq:li_general_eigenVI}
\end{equation}
where $1\leq r\leq K^D$.
When $r=1$, we can write $\rho=\theta\theta^\top$ with $\|\theta\|_2=1$, recovering the original EigenVI formulation in~\eqref{eq:cai_eigen}.
For general $r$, the variational parameter becomes a positive semidefinite (PSD) matrix $\rho\in\mathbb{R}^{K^D\times K^D}$ with unit trace.
Such a matrix is called a \emph{density operator} in quantum information theory.
Thus, EigenVI corresponds to the rank-one case of~\eqref{eq:li_general_eigenVI}.

The density-operator formulation allows multiple eigen-directions to contribute to the variational distribution.
In particular, when the minimum eigenvalue of $M$ has multiplicity $r$, any density operator supported on the corresponding eigenspace is optimal for~\eqref{eq:li_general_eigenVI}.
This includes mixed states that combine all $r$ orthonormal eigenvectors, extending the single-eigenvector representation of EigenVI.

We remark that our recent work~\citep{li2026approximation} explains a connection between~\eqref{eq:li_general_eigenVI} and Fisher divergence minimization.
Specifically, the minimum loss of~\eqref{eq:li_general_eigenVI} is upper bounded by that of EigenVI and lower bounded by $D_F(q_{\rho^\star}\Vert p)$, where $\rho^\star$ is an optimal solution to~\eqref{eq:li_general_eigenVI} and $q_{\rho^\star}(x):=\Phi(x)^\top\rho^\star\Phi(x)$ is the corresponding variational distribution.
It implies that minimizing~\eqref{eq:li_general_eigenVI} is \emph{not} equivalent to directly minimizing the Fisher divergence, but remains a reasonable surrogate loss, as its minimum is sandwiched between the Fisher divergence of the resulting variational distribution and the minimum Fisher loss of EigenVI.

\subsection{Parameterizing the Density Operator $\rho$ with QTNs}
\label{sec:qtn_parameterization}

In \model{}, we parameterize the density operator $\rho$ in~\eqref{eq:li_general_eigenVI} using a specific QTN architecture.
The corresponding graphical representation is given by
\begingroup
\setlength{\abovedisplayskip}{4pt}
\setlength{\belowdisplayskip}{4pt}
\begin{equation}
\label{eq:rho_def}
\rho =
\vcenter{\hbox{%
    \includegraphics[
        width=0.48\textwidth
    ]{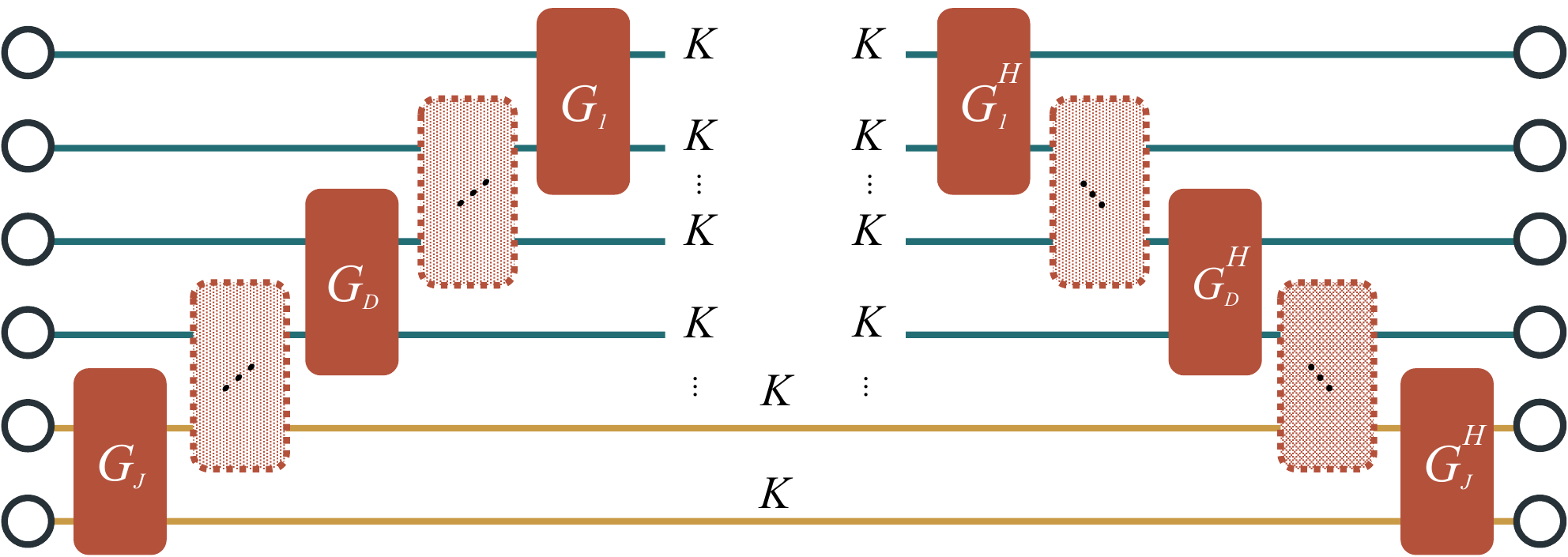}%
}}.
\end{equation}
\endgroup
As shown in~\eqref{eq:rho_def}, the QTN consists of learnable gate tensors $G_j\in\mathbb{C}^{K^L\times K^L}$, for $j=1,\ldots,J$, and fixed initial-state tensors.
Here, $K$ denotes the wire bond dimension, determined by the \emph{basis size}, while $L$, referred to as the \emph{local window size}, denotes the number of wires acted upon by each gate tensor.
Each entry of the initial-state tensors is set to $e^{i\pi/4}/\sqrt K$, yielding a normalized uniform superposition state.

Unlike general QTNs, the model in~\eqref{eq:rho_def} has three distinctive features.
First, the gate tensors are paired with their complex conjugates in a mirrored structure, ensuring that the resulting density operator is PSD.
Second, the gate tensors are arranged in a staircase pattern, where each gate shares one wire with its neighbors.
This structure is known to be equivalent to an MPO representation~\citep{ran2020encoding}.
As we show later, the staircase structure also enables efficient computation during both training and inference.
Third, the wires in the QTN are divided into two types, distinguished by color: variable wires and ancilla wires.
The variable wires correspond to the indices of the density operator $\rho$, while the ancilla wires enable the representation of mixed states according to the purification theorem~\citep{nielsen2010quantum}.
The number of variable wires is equal to the data dimension $D$, while the number of ancilla wires is a hyperparameter that controls the maximum rank of the density operator.

\subsection{Localizing the Loss-Induced Matrix $M$}
\label{sec:local_qtn}

In EigenVI, the integral in~\eqref{eq:def_M} is estimated by importance sampling using a $D$-dimensional uniform proposal.
However, the estimation variance can grow exponentially with $D$, requiring exponentially more samples to maintain the same accuracy.
To address this challenge, we exploit the local dependency structure of the target distribution, inspired by local interaction models in statistical physics, such as the Ising model.
Under this assumption, we show that $M$ can be expressed as a sum of local terms, each involving a bounded number of variables independent of $D$.
This decomposition allows importance sampling to be applied to each local term separately.

Specifically, let $s_d(x):=\partial_{x_d}\log p(x)$ denote the $d$-th component of the target score.
We assume a nearest-neighbor dependency structure, where $s_d(x)$ depends only on the variables in $\mathcal{N}_d:=\{d-1,d,d+1\}\cap[D]$.
Let $\Phi_{\mathcal{N}_d}(x_{\mathcal{N}_d})$ denote the tensor-product basis restricted to $\mathcal{N}_d$.
We define the local matrix
\begin{equation}
\begin{aligned}
M^{(d)}
&:= \int
\Big(
2\partial_{x_d}\Phi_{\mathcal{N}_d}(x_{\mathcal{N}_d})
- s_d(x_{\mathcal{N}_d})\Phi_{\mathcal{N}_d}(x_{\mathcal{N}_d})
\Big)\\
&\qquad\cdot
\Big(
2\partial_{x_d}\Phi_{\mathcal{N}_d}(x_{\mathcal{N}_d})
- s_d(x_{\mathcal{N}_d})\Phi_{\mathcal{N}_d}(x_{\mathcal{N}_d})
\Big)^\top
\,dx_{\mathcal{N}_d}.
\end{aligned}
\label{eq:local_Md}
\end{equation}

The loss-induced matrix $M$ can then be constructed as
\begin{equation}
\begin{aligned}
M
={}& M^{(1)}\otimes I_K^{\otimes(D-2)}+\sum_{d=2}^{D-1}
I_K^{\otimes(d-2)}
\otimes M^{(d)}
\otimes I_K^{\otimes(D-d-1)}+I_K^{\otimes(D-2)}\otimes M^{(D)},
\end{aligned}
\label{eq:local_M}
\end{equation}
where $I_K$ denotes the $K\times K$ identity matrix.
Each local matrix $M^{(d)}$ has dimension at most $K^3\times K^3$, independent of $D$.
Thus, each $M^{(d)}$ can be estimated separately through importance sampling over at most three variables, avoiding integration over the full $D$-dimensional space.

\begin{figure}[t]
    \centering
    {\scriptsize
    \[
    \operatorname{tr}\left(
        \rho_{\mathcal{N}_1}M^{(1)}
    \right)
    =
    \vcenter{\hbox{%
        \includegraphics[width=0.31\textwidth]
        {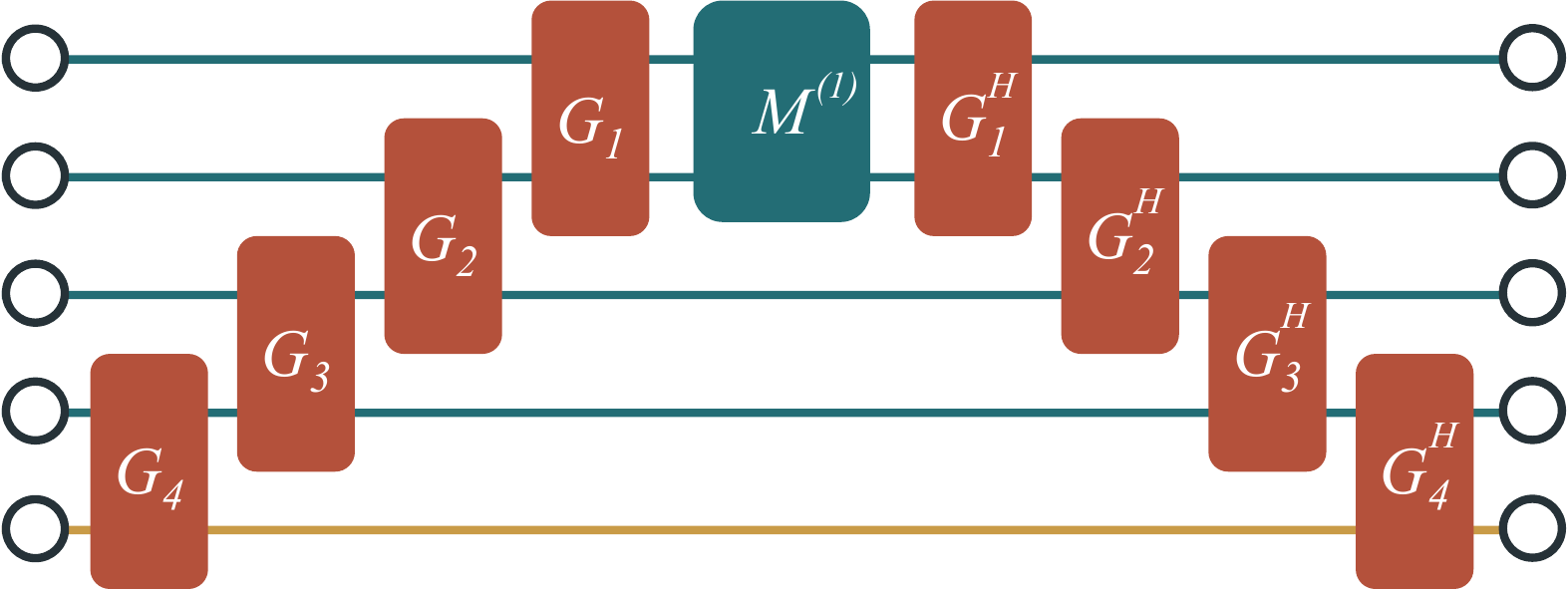}%
    }}
    \qquad
    \operatorname{tr}\left(
        \rho_{\mathcal{N}_2}M^{(2)}
    \right)
    =
    \vcenter{\hbox{%
        \includegraphics[width=0.31\textwidth]
        {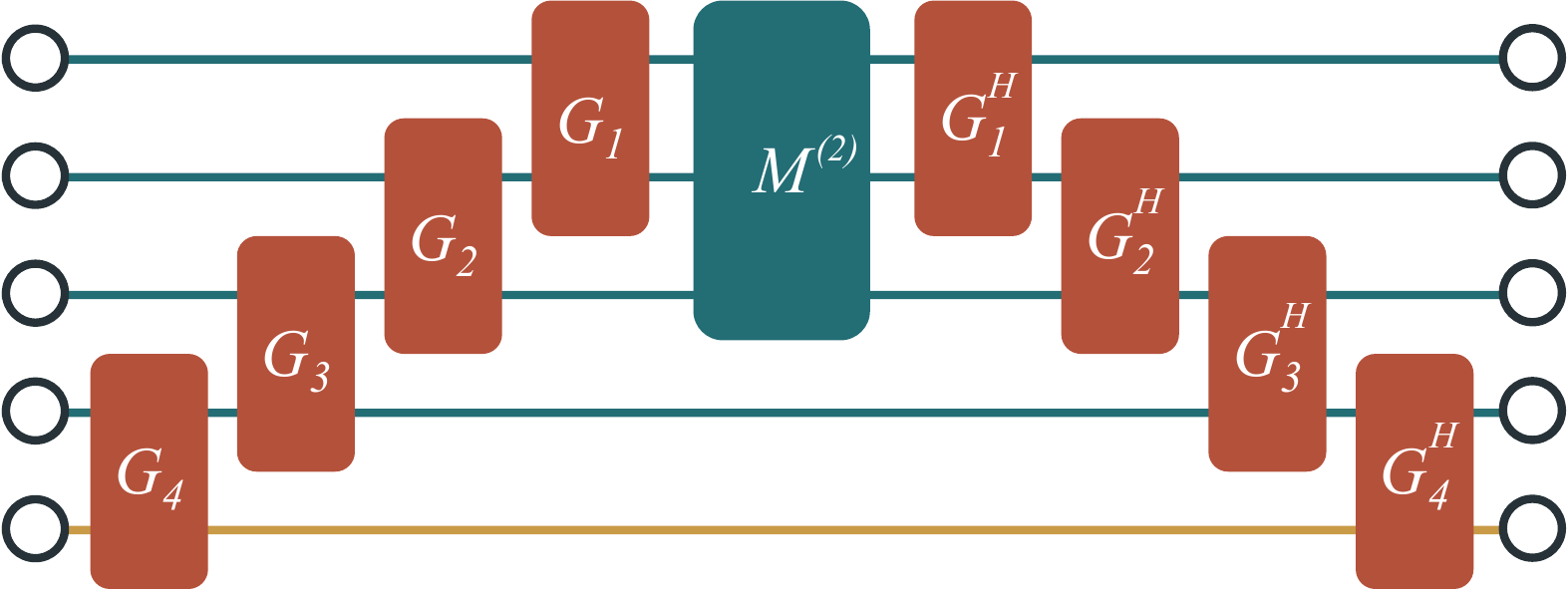}%
    }}
    \]
    }

    \caption{Illustration of tensor-network contractions for
$\operatorname{tr}(\rho_{\mathcal{N}_1}M^{(1)})$ and
$\operatorname{tr}(\rho_{\mathcal{N}_2}M^{(2)})$
in~\eqref{eq:local_loss}, with $D=4$, $L=2$, and one ancilla wire.}

    \label{fig:training}
\end{figure}

Substituting~\eqref{eq:local_M} into the loss in~\eqref{eq:li_general_eigenVI}, we obtain
\begin{equation}
    \operatorname{tr}(\rho M)
    =
    \sum_{d=1}^{D}
    \operatorname{tr}\left(\rho_{\mathcal{N}_d}M^{(d)}\right),
    \label{eq:local_loss}
\end{equation}
where $\rho_{\mathcal{N}_d}:=\operatorname{tr}_{[D]\setminus\mathcal{N}_d}(\rho)$ denotes the reduced density operator obtained by taking the partial trace~\citep{nielsen2010quantum} over variables outside $\mathcal{N}_d$.
Consequently, each local loss term involves only matrices of dimension at most $K^3\times K^3$.
Figure~\ref{fig:training} illustrates how to compute two local loss terms in~\eqref{eq:local_loss} using tensor-network contractions.
The QTN structure enables efficient computation of these reduced density operators without explicitly constructing the full density operator $\rho$.

We remark that the nearest-neighbor assumption does not necessarily hold for arbitrary target distributions.
In general, the construction in~\eqref{eq:local_M} provides an approximation to the original loss-induced matrix defined in~\eqref{eq:def_M}.
However, as established in~\citep{li2026approximation}, this construction is exact when the target distribution is a path-graph Markov random field.

\section{Algorithm}
\label{algorithm}

In this section, we present the algorithms for efficient training and inference with \model{}.


\subsection{Training}

For each $d$, let $\widetilde{M}^{(d)}$ denote an estimate of $M^{(d)}$ obtained through importance sampling.
Following~\eqref{eq:local_loss}, we define the training loss of \model{} as
$
    \mathcal{L}(\{G_j\}_{j=1}^J)
    :=
    \sum_{d=1}^{D}
    \operatorname{tr}\left(
        \rho_{\mathcal{N}_d}\widetilde{M}^{(d)}
    \right),
$
where $\rho_{\mathcal{N}_d}$ is parameterized by the gate tensors $\{G_j\}_{j=1}^J$.

Training proceeds as follows.
First, the gate tensors are initialized using Haar-Gaussian-distributed unitary matrices.
Next, the loss is evaluated through tensor-network contractions, proceeding wire by wire from top to bottom.
Finally, the gate tensors are updated by gradient descent on the Stiefel manifold until convergence.

This training procedure offers two advantages.
First, for fixed $K$ and $L$, the contractions require only polynomial computational time and memory in $D$.
Second, Stiefel manifold optimization preserves the unitary constraints of the gate tensors, ensuring that $\rho$ remains a valid density operator without explicit normalization.
Implementation details are provided in our Python code, with a complexity analysis in Appendix~\ref{app:complexity}.

\subsection{Inference}
We consider four inference tasks for \model{}: density queries, marginalization, conditionalization, and sampling.
All these tasks can be performed efficiently through tensor-network contractions.


\paragraph{Density queries and marginalization.}
After training, density queries and marginalization can be performed through tensor-network contractions, based on $q(x)=\Phi(x)^\top\rho\Phi(x)$.
As an illustrative example, consider $x=(x_1,x_2,x_3,x_4)$ with $D=4$, $L=2$, and one ancilla wire.
Let $m(x_d):=\phi(x_d)\phi(x_d)^\top$ for each $d\in[D]$.
The joint density $q(x)$ and the marginal density $q(x_2,x_4)$, obtained by integrating out $x_1$ and $x_3$, can be computed as follows:
\begin{equation}
    q(x)
    =
    \vcenter{\hbox{%
        \includegraphics[width=0.34\textwidth]
        {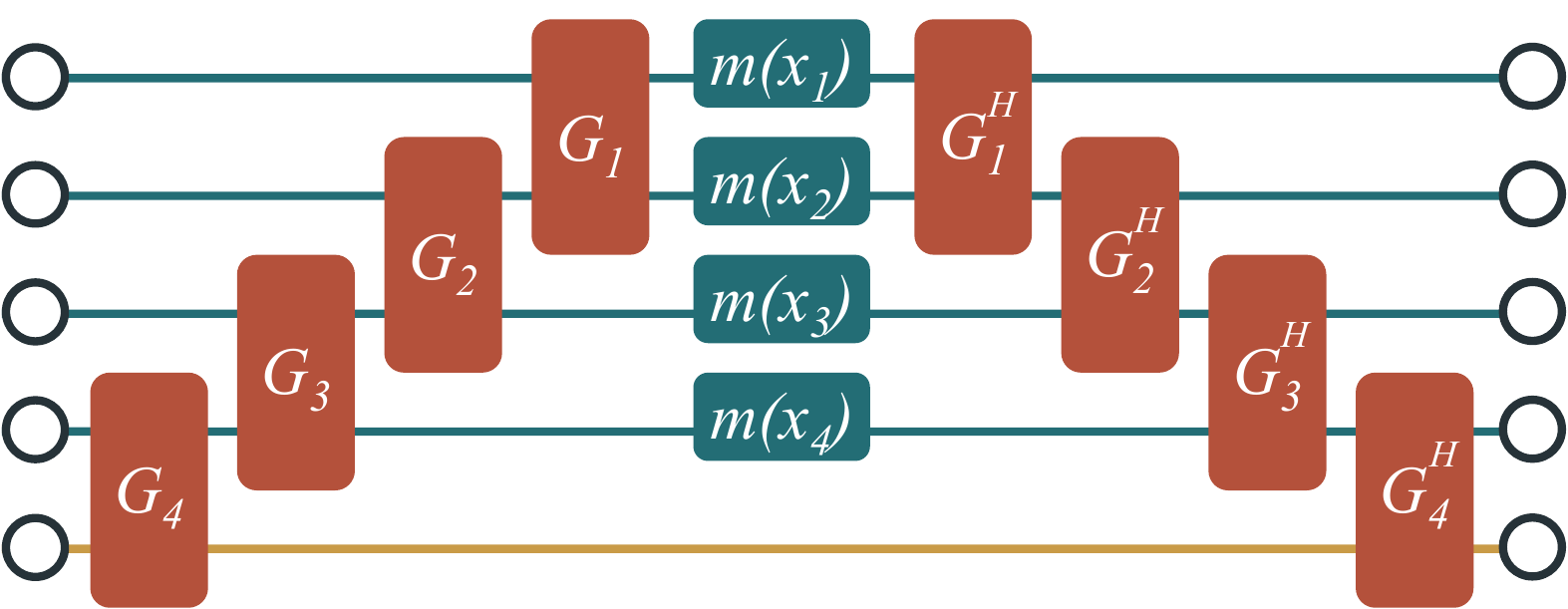}%
    }},
    \quad
    q(x_2,x_4)
    =
    \vcenter{\hbox{%
        \includegraphics[width=0.34\textwidth]
        {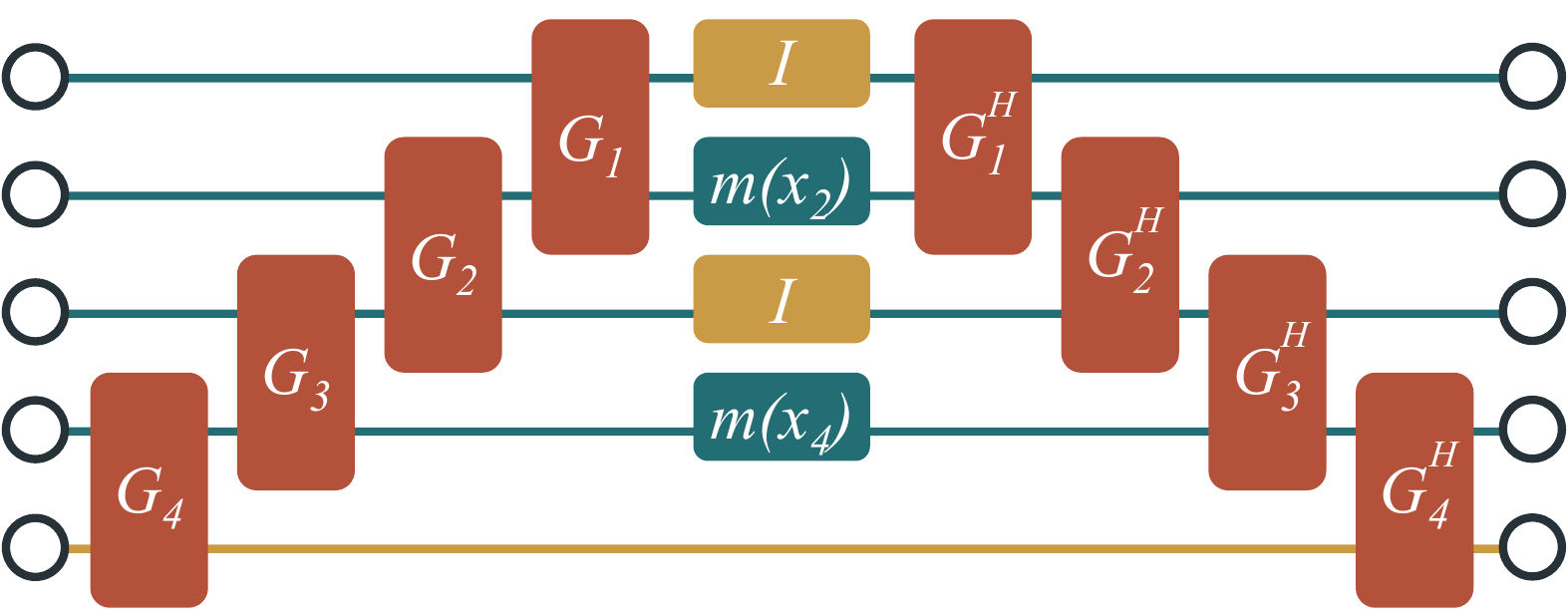}%
    }}.
\end{equation}
Here, marginalization is performed by replacing $m(x_d)$ with the identity matrix $I$ for each marginalized variable $x_d$, since the orthonormality of the basis functions ensures that
$\int m(x_d)\,dx_d
=\int\phi(x_d)\phi(x_d)^\top\,dx_d=I$.

\paragraph{Conditional density.}
Conditional densities can be calculated as ratios of tensor-network contractions.
Under the same setting with $D=4$, $L=2$, and one ancilla wire, consider the conditional density of $(x_1,x_3)$ given $(x_2,x_4)=(z_2,z_4)$:
$
    q(x_1,x_3\mid z_2,z_4)
    =
    \frac{q(x_1,z_2,x_3,z_4)}{q(z_2,z_4)}
$.
The numerator is obtained by fixing the local matrices at $(x_1,z_2,x_3,z_4)$, while the denominator is obtained by replacing the matrices at sites $1$ and $3$ with identity matrices.
The corresponding tensor-network contractions are illustrated as follows:
\begin{equation}
    q(x_1,x_3\mid z_2,z_4)
    =
    \vcenter{\hbox{%
        \includegraphics[width=0.33\textwidth]
        {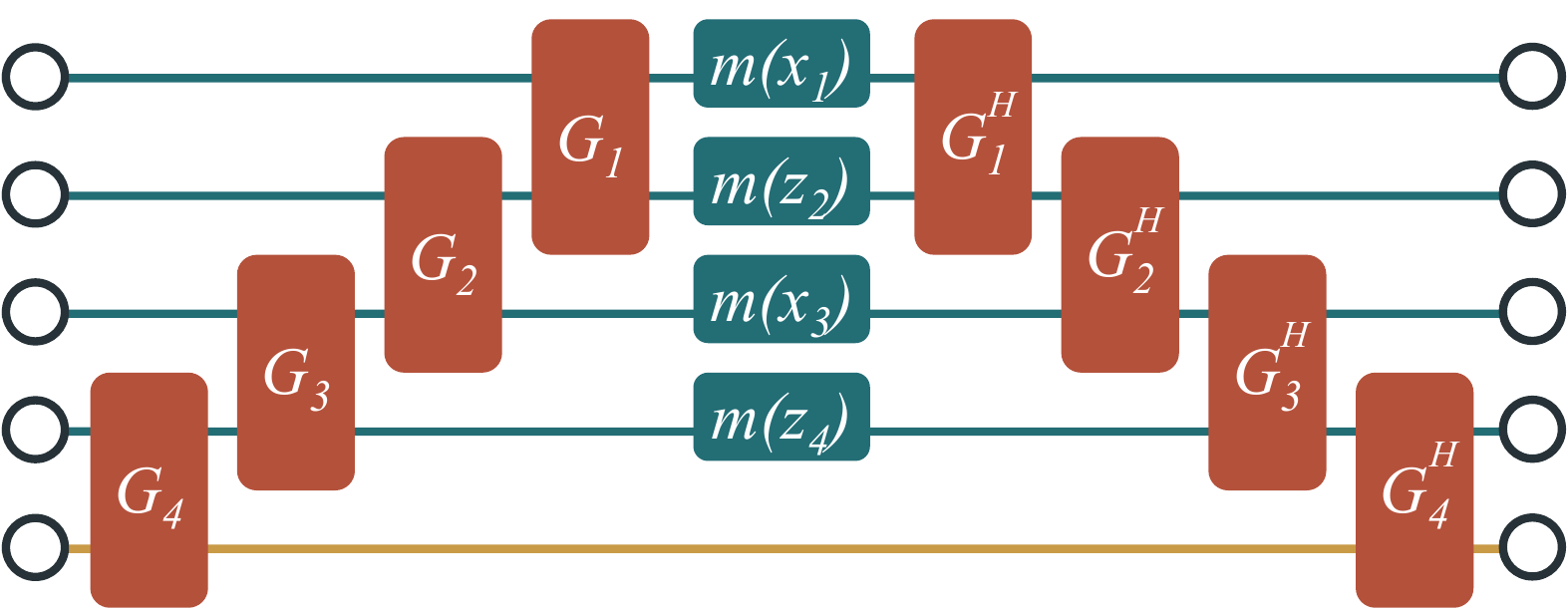}%
    }}
    \Big/
    \vcenter{\hbox{%
        \includegraphics[width=0.33\textwidth]
        {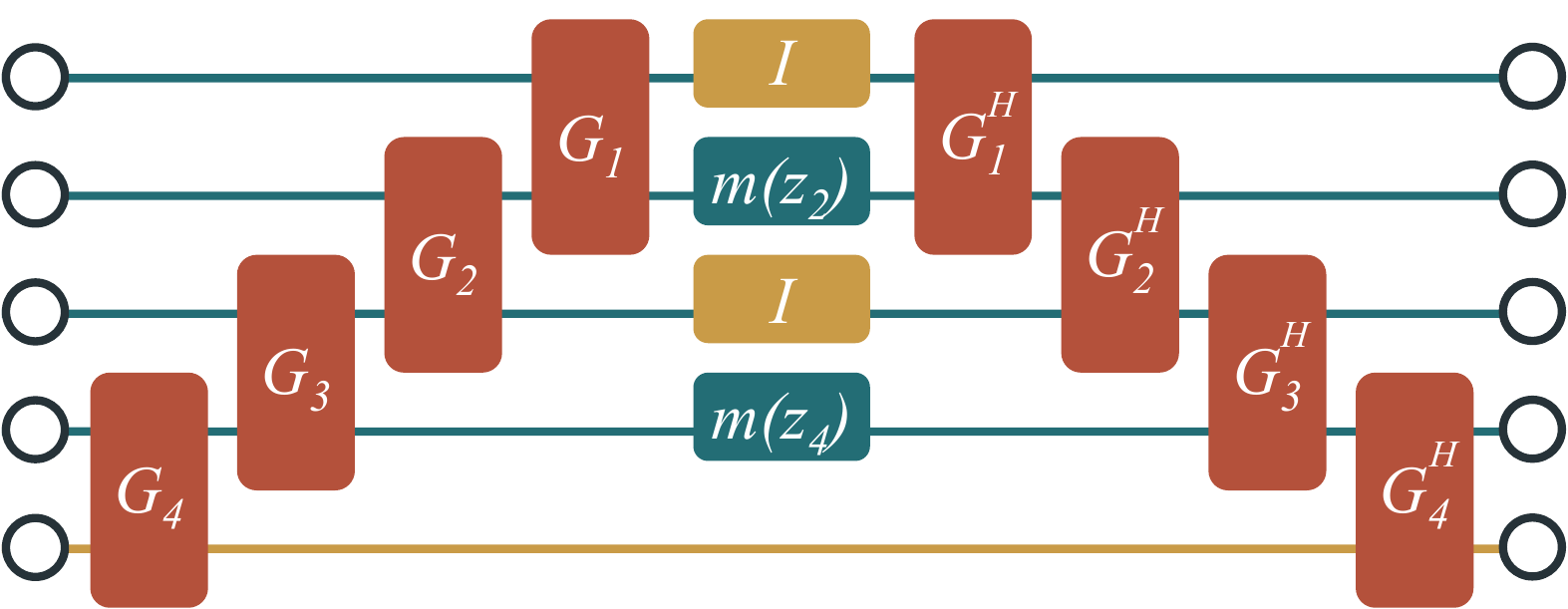}%
    }}.
\end{equation}

\paragraph{Sampling.}
Sampling follows the probabilistic chain rule:
$
    q(x_1,\ldots,x_D)
    =
    q(x_1)
    \prod_{d=2}^{D}
    q(x_d\mid x_{<d}),
$
where $x_{<d}:=(x_1,\ldots,x_{d-1})$.
Following EigenVI~\citep{cai2024eigenvi}, \model{} generates samples sequentially using this factorization.
At each step $d$, the previously sampled variables $x_{<d}$ are fixed, while the future variables $x_{>d}$ are marginalized out by replacing their local matrices with identity matrices.
The resulting tensor-network contractions yield the one-dimensional conditional density
$
    q(x_d\mid x_{<d})
    =
    \frac{q(x_{\le d})}{q(x_{<d})}.
$
We then numerically integrate this density to obtain its conditional cumulative distribution function (CDF), draw $u\sim\mathrm{Uniform}(0,1)$, and obtain $x_d$ by inverting the CDF.
Repeating this procedure for $d=1,\ldots,D$ produces a sample from $q$.

Density queries, marginalization, conditional density evaluation, and sampling all rely on the same tensor-network contraction framework.
A detailed complexity analysis is provided in Appendix~\ref{app:complexity}.

\section{Experiments}
\label{experiments}

\newcommand{\meanstd}[2]{#1\,{\scriptstyle \pm #2}}

\begin{table*}[t]
\centering
\caption{
Forward KL divergence (mean $\pm$ standard deviation over five runs; lower is better).
The best mean in each row is highlighted in \textbf{bold}.
For \model{}, $E$ denotes the number of ancilla wires.
\textcolor{gray}{OOM} indicates out-of-memory, and ``--'' indicates unstable runs.
}
\label{tab:synthetic_dimension_scaling}

\begingroup
\setlength{\tabcolsep}{3.5pt}
\renewcommand{\arraystretch}{1.12}

\resizebox{\textwidth}{!}{%
\begin{tabular}{
    ll
    ccccc
    >{\columncolor{quanviblue}}c
    >{\columncolor{quanviblue}}c
}
\toprule

\multirow{2}{*}{$D$}
& \multirow{2}{*}{Target}
& \multicolumn{5}{c}{Baselines}
& \multicolumn{2}{c}{\textbf{\model{}}} \\

\cmidrule(lr){3-7}
\cmidrule(l){8-9}

&
& \shortstack{GSM\\[-1pt]\scriptsize\citep{modi2023variational}}
& \shortstack{BAM\\[-1pt]\scriptsize\citep{caibatch}}
& \shortstack{ADVI\\[-1pt]\scriptsize\citep{kucukelbir2017automatic}}
& \shortstack{EigenVI\\[-1pt]\scriptsize\citep{cai2024eigenvi}}
& \shortstack{MoG\\[-1pt]\scriptsize\citep{petit2025variational}}
& $E=0$
& $E=2$ \\

\midrule

\multirow{5}{*}{$5$}
& Gaussian
& $\meanstd{0.9520}{1.3282}$
& $\bm{\le 1{\times}10^{-2}}$
& $\bm{\le 1{\times}10^{-2}}$
& $\bm{\le 1{\times}10^{-2}}$
& $\meanstd{0.0152}{0.0009}$
& $\bm{\le 1{\times}10^{-2}}$
& $\bm{\le 1{\times}10^{-2}}$ \\

& X-shape
& $\meanstd{2.2494}{0.4961}$
& $\meanstd{3.0904}{0.2373}$
& $\meanstd{1.6816}{0.0716}$
& $\meanstd{0.4007}{0.0743}$
& $\meanstd{1.3039}{0.0269}$
& $\bm{\meanstd{0.2613}{0.0035}}$
& $\meanstd{0.2661}{0.0054}$ \\

& GMM3
& $\meanstd{1.6469}{0.5895}$
& $\meanstd{2.4417}{0.2273}$
& $\meanstd{1.1967}{0.0313}$
& $\meanstd{0.4454}{0.0613}$
& $\meanstd{0.4436}{0.0617}$
& $\meanstd{0.1608}{0.0112}$
& $\bm{\meanstd{0.1598}{0.0042}}$ \\

& Funnel
& $\meanstd{3.3535}{0.5241}$
& $\meanstd{4.8814}{0.2082}$
& $\meanstd{2.8012}{0.0653}$
& $\meanstd{1.3435}{0.1065}$
& $\bm{\meanstd{0.8183}{0.0742}}$
& $\meanstd{4.3944}{1.8299}$
& $\meanstd{2.6412}{0.4579}$ \\

& Ring
& $\meanstd{30.7895}{8.8223}$
& $\meanstd{39.8048}{2.5664}$
& $\meanstd{2.1094}{0.0221}$
& $\meanstd{1.7526}{0.1454}$
& $\meanstd{2.0528}{0.2233}$
& $\bm{\meanstd{0.8861}{0.0048}}$
& $\meanstd{0.8913}{0.0144}$ \\

\midrule

\multirow{5}{*}{$10$}
& Gaussian
& $\bm{\le 1{\times}10^{-2}}$
& $\bm{\le 1{\times}10^{-2}}$
& $\bm{\le 1{\times}10^{-2}}$
& \textcolor{gray}{OOM}
& $\meanstd{0.0814}{0.0024}$
& $\bm{\le 1{\times}10^{-2}}$
& $\bm{\le 1{\times}10^{-2}}$ \\

& X-shape
& $\meanstd{16.0632}{5.3750}$
& $\meanstd{24.9337}{1.6719}$
& $\meanstd{5.8356}{0.0990}$
& \textcolor{gray}{OOM}
& $\meanstd{6.5115}{0.1666}$
& $\meanstd{3.2554}{0.0670}$
& $\bm{\meanstd{3.1928}{0.0655}}$ \\

& GMM3
& $\meanstd{7.0765}{1.1840}$
& $\meanstd{24.0342}{1.3684}$
& $\meanstd{5.4414}{0.0758}$
& \textcolor{gray}{OOM}
& $\meanstd{5.1997}{0.3525}$
& $\bm{\meanstd{3.1225}{0.0504}}$
& $\meanstd{3.1551}{0.0551}$ \\

& Funnel
& \textcolor{gray}{--}
& $\meanstd{29.5528}{4.4531}$
& $\meanstd{7.0246}{0.0623}$
& \textcolor{gray}{OOM}
& $\bm{\meanstd{5.0368}{0.3313}}$
& $\meanstd{9.0460}{0.8675}$
& $\meanstd{8.8288}{1.0972}$ \\

& Ring
& $\meanstd{9.1494}{1.9204}$
& $\meanstd{64.7096}{1.8813}$
& $\meanstd{6.3256}{0.1146}$
& \textcolor{gray}{OOM}
& $\meanstd{7.7603}{0.4014}$
& $\meanstd{3.8400}{0.0213}$
& $\bm{\meanstd{3.8302}{0.0271}}$ \\

\midrule

\multirow{5}{*}{$20$}
& Gaussian
& $\bm{\le 1{\times}10^{-2}}$
& $\bm{\le 1{\times}10^{-2}}$
& $\meanstd{0.0124}{0.0035}$
& \textcolor{gray}{OOM}
& $\meanstd{0.2884}{0.0088}$
& $\bm{\le 1{\times}10^{-2}}$
& $\bm{\le 1{\times}10^{-2}}$ \\

& X-shape
& $\meanstd{14.3388}{2.1860}$
& $\meanstd{38.0141}{2.2533}$
& $\meanstd{8.4358}{0.1412}$
& \textcolor{gray}{OOM}
& $\meanstd{9.9434}{0.2302}$
& $\bm{\meanstd{8.1714}{0.8125}}$
& $\meanstd{8.9266}{0.5087}$ \\

& GMM3
& $\meanstd{11.4537}{0.9701}$
& $\meanstd{38.2367}{0.5113}$
& $\meanstd{8.1050}{0.1880}$
& \textcolor{gray}{OOM}
& $\meanstd{8.4863}{0.4095}$
& $\bm{\meanstd{7.4542}{1.4717}}$
& $\meanstd{7.7778}{1.1659}$ \\

& Funnel
& \textcolor{gray}{--}
& $\meanstd{41.6810}{3.9465}$
& $\meanstd{9.8232}{0.1614}$
& \textcolor{gray}{OOM}
& $\bm{\meanstd{7.6320}{0.2791}}$
& $\meanstd{13.1389}{0.9869}$
& $\meanstd{13.4216}{0.9977}$ \\

& Ring
& $\meanstd{13.5824}{0.2441}$
& $\meanstd{77.6129}{5.2570}$
& $\bm{\meanstd{8.9786}{0.1421}}$
& \textcolor{gray}{OOM}
& $\meanstd{10.6938}{0.3243}$
& $\meanstd{9.1621}{1.5400}$
& $\meanstd{9.0459}{1.0407}$ \\

\midrule

\multirow{5}{*}{$100$}
& Gaussian
& $\bm{\le 1{\times}10^{-2}}$
& $\bm{\le 1{\times}10^{-2}}$
& $\meanstd{0.1308}{0.0020}$
& \textcolor{gray}{OOM}
& $\meanstd{2.2290}{0.0119}$
& $\bm{\le 1{\times}10^{-2}}$
& $\bm{\le 1{\times}10^{-2}}$ \\

& X-shape
& $\meanstd{75.3810}{2.1274}$
& $\meanstd{186.0990}{4.2477}$
& $\meanstd{47.2649}{0.2927}$
& \textcolor{gray}{OOM}
& $\meanstd{72.7630}{6.0097}$
& $\meanstd{42.2971}{1.2528}$
& $\bm{\meanstd{40.5473}{4.0462}}$ \\

& GMM3
& $\meanstd{73.5309}{3.1276}$
& $\meanstd{187.8098}{4.1833}$
& $\meanstd{47.1609}{0.2688}$
& \textcolor{gray}{OOM}
& $\meanstd{68.3951}{7.4535}$
& $\bm{\meanstd{43.0097}{2.3348}}$
& $\meanstd{43.0798}{2.9883}$ \\

& Funnel
& $\meanstd{85.7464}{3.1931}$
& $\meanstd{188.2434}{3.0382}$
& $\meanstd{48.8944}{0.2271}$
& \textcolor{gray}{OOM}
& $\meanstd{57.0316}{0.4351}$
& $\bm{\meanstd{46.4010}{3.1020}}$
& $\meanstd{47.0005}{3.8384}$ \\

& Ring
& $\meanstd{73.7223}{2.1166}$
& $\meanstd{224.9920}{4.1197}$
& $\meanstd{48.1236}{0.2075}$
& \textcolor{gray}{OOM}
& $\meanstd{76.9241}{2.1116}$
& $\meanstd{43.7935}{1.0249}$
& $\bm{\meanstd{43.1974}{1.5993}}$ \\

\bottomrule
\end{tabular}%
}
\endgroup
\end{table*}

We evaluate \model{} through three sets of experiments. First, we assess its performance and scalability on synthetic distributions across different dimensions and target characteristics. Second, we evaluate posterior approximation on \textsc{PosteriorDB}~\citep{magnusson2024posteriordb}. 
Finally, we conduct ablation studies on the window size $L$, basis size $K$, and number of ancilla wires $E$.

\subsection{Synthetic target distributions}
\label{sec:exp_synthetic}

\paragraph{Target distributions and evaluation metric.}
We consider five synthetic target distributions:
Gaussian, X-shape, GMM3, Funnel, and Ring, illustrated in Figure~\ref{fig:synthetic-targets} of the Appendix.
The non-Gaussian targets are extended from two to $D$ dimensions through nonlinear Markov-chain augmentation, while the Gaussian target is constructed directly in $D$ dimensions with local correlations.
Details are provided in Appendix~\ref{app:high-dim-transform}.
Following EigenVI, we evaluate approximation quality using the forward KL divergence $\mathrm{KL}(p\Vert q)$, estimated with $10^4$ Monte Carlo samples drawn from $p$.

\paragraph{Baselines.}
We compare \model{} with ADVI~\citep{kucukelbir2017automatic},
BAM~\citep{caibatch}, GSM~\citep{modi2023variational},
MoG~\citep{petit2025variational}, and EigenVI~\citep{cai2024eigenvi}.
For \model{}, we consider $E=0$ and $E=2$, corresponding to models without and with ancilla wires, respectively.
EigenVI is evaluated only for $D\le 5$, as its global $K^D\times K^D$ eigendecomposition becomes computationally infeasible in higher dimensions.
We also evaluate normalizing flows, but report their results separately in Appendix~\ref{app:nf} due to unstable performance across targets.
Hyperparameter settings are provided in Appendix~\ref{syn_para}.

\paragraph{Discussion.}
Table~\ref{tab:synthetic_dimension_scaling} shows that \model{} achieves competitive accuracy with EigenVI in low dimensions while scaling to higher dimensions where EigenVI becomes computationally infeasible. At $D=5$, \model{} matches or outperforms EigenVI on most targets. In contrast, EigenVI cannot scale to higher dimensions due to its global $K^D\times K^D$ eigendecomposition.

On Gaussian targets, \model{} consistently achieves near-zero KL divergence across all tested dimensions, demonstrating its effectiveness on distributions that satisfy the local dependency assumption. For non-Gaussian targets, \model{} achieves the best or near-best results on X-shape, GMM3, and Ring in most settings. The main exception is Funnel, where the default configuration is less competitive. As shown in Section~\ref{sec:exp_ablation}, increasing the local window size substantially improves approximation accuracy on Funnel, at the cost of higher computational complexity. This illustrates the trade-off between expressiveness and computational efficiency.

Figure~\ref{fig:xshape_d20} compares samples generated by different methods, projected onto the first two dimensions of the $D=20$ X-shape target. While GSM and the mixture-based baselines fail to recover the characteristic X-shaped geometry, \model{} more closely reproduces the target distribution.

\subsection{Bayesian posterior approximation}
\label{sec:exp_posterior}

Next, we evaluate \model{} on Bayesian posterior approximation tasks from \textsc{PosteriorDB}~\citep{magnusson2024posteriordb}. Since the normalized target densities are unavailable, we follow EigenVI and evaluate approximation quality using the forward Fisher divergence, estimated from reference samples provided by the dataset.

Table~\ref{tab:posterior_benchmark} presents the numerical results. Across three posterior benchmarks, \model{} achieves accuracy comparable to EigenVI while substantially outperforming the other baselines. In particular, \model{} achieves the lowest Fisher divergence on \texttt{hmm\_bball\_0}. 
Figure~\ref{fig:hmm_bball_0} compares the one-dimensional marginals of \model{} and EigenVI against reference samples on this task, demonstrating that \model{} more accurately captures the non-Gaussian structure of the posterior.

\begin{table*}[t]
\centering
\caption{
Bayesian posterior approximation results measured by forward Fisher divergence
(mean $\pm$ standard deviation over four runs; lower is better).
Best mean values are highlighted in \textbf{bold}.
For \model{}, $E$ denotes the number of ancilla wires.
\texttt{bball} denotes \texttt{hmm\_bball\_0}, and ``--'' indicates unstable runs.
}
\label{tab:posterior_benchmark}

\begingroup
\setlength{\tabcolsep}{5pt}
\renewcommand{\arraystretch}{1.3}

\resizebox{\textwidth}{!}{%
\begin{tabular}{
    lccccccc
    >{\columncolor{quanviblue}}c
    >{\columncolor{quanviblue}}c
}
\toprule
Model & $D$ & GSM & BAM & ADVI & MoG & NF~\cite{rezende2015variational} & EigenVI
& \model{} ($E=0$) & \model{} ($E=2$) \\
\midrule

\texttt{gpregr} & 3
& $\meanstd{1.2107}{0.0345}$
& $\meanstd{1.1962}{0.0293}$
& $\meanstd{1.1509}{0.0177}$
& $\meanstd{7.6146}{3.9904}$
& $\meanstd{3.3739}{0.4973}$
& $\bm{\meanstd{0.0828}{0.0681}}$
& $\meanstd{0.2180}{0.0463}$
& $\meanstd{0.2330}{0.0369}$ \\

\addlinespace[3pt]

\texttt{hmm} & 4
& $\meanstd{148.1225}{2.5882}$
& $\meanstd{139.7349}{3.9980}$
& $\meanstd{168.9795}{32.6976}$
& \textcolor{gray}{--}
& \textcolor{gray}{--}
& $\bm{\meanstd{4.0295}{1.5069}}$
& $\meanstd{5.0783}{0.6214}$
& $\meanstd{4.8047}{0.4491}$ \\

\addlinespace[3pt]

\texttt{bball} & 6
& $\meanstd{749.1771}{2.5298}$
& $\meanstd{19.9691}{0.6849}$
& $\meanstd{746.7151}{2.0978}$
& \textcolor{gray}{--}
& \textcolor{gray}{--}
& $\meanstd{18.5057}{3.0322}$
& $\meanstd{13.7126}{0.5198}$
& $\bm{\meanstd{13.4460}{0.3597}}$ \\

\bottomrule
\end{tabular}%
}
\endgroup
\end{table*}


\subsection{Ablation study}
\label{sec:exp_ablation}
We further conduct ablation studies to examine how key design
choices affect approximation quality and the trade-off between
expressiveness and computational cost.

\paragraph{Local window size \(L\).}
\label{abla_l}

\providecommand{\klstd}[2]{%
    \ensuremath{#1\,{\scriptstyle\pm #2}}%
}

\begin{table}[t]
\centering
\caption{
Effect of local window size $L$ on forward KL divergence
(mean $\pm$ standard deviation over five runs; lower is better).
The best mean in each row is highlighted in \textbf{bold}.
``--'' indicates unavailable results.
}
\label{tab:ablation_local_window}

\vspace{4pt}

\begingroup
\footnotesize
\setlength{\tabcolsep}{3pt}
\renewcommand{\arraystretch}{1.15}

\begin{tabular}{clccccc}
\toprule
$D$ & Target
& $L=1$ & $L=2$ & $L=3$ & $L=4$ & $L=5$ \\
\midrule

\multirow{4}{*}{5}
& X-shape
& \klstd{6.773}{0.040}
& \klstd{0.265}{0.007}
& \klstd{0.248}{0.005}
& \klstd{0.255}{0.008}
& $\bm{\klstd{0.246}{0.012}}$ \\

& GMM3
& \klstd{9.266}{4.613}
& \klstd{0.160}{0.004}
& \klstd{0.156}{0.005}
& $\bm{\klstd{0.155}{0.007}}$
& $\bm{\klstd{0.155}{0.005}}$ \\

& Funnel
& \klstd{9.299}{2.148}
& \klstd{2.838}{1.262}
& \klstd{1.606}{0.212}
& $\bm{\klstd{1.388}{0.093}}$
& \klstd{1.481}{0.265} \\

& Ring
& \klstd{11.050}{3.082}
& $\bm{\klstd{0.879}{0.018}}$
& \klstd{0.891}{0.025}
& \klstd{0.919}{0.027}
& \klstd{0.893}{0.024} \\

\midrule

\multirow{4}{*}{10}
& X-shape
& \klstd{16.203}{1.615}
& \klstd{3.212}{0.069}
& \klstd{2.956}{0.053}
& $\bm{\klstd{2.859}{0.053}}$
& \klstd{2.869}{0.060} \\

& GMM3
& \klstd{19.865}{5.100}
& \klstd{3.146}{0.065}
& \klstd{2.905}{0.049}
& \klstd{2.734}{0.068}
& $\bm{\klstd{2.728}{0.097}}$ \\

& Funnel
& \klstd{18.983}{3.115}
& \klstd{7.929}{1.195}
& \klstd{6.367}{0.435}
& \klstd{4.383}{0.527}
& $\bm{\klstd{2.812}{0.050}}$ \\

& Ring
& \klstd{20.669}{3.089}
& \klstd{3.822}{0.047}
& \klstd{3.580}{0.074}
& \klstd{3.450}{0.060}
& $\bm{\klstd{3.343}{0.064}}$ \\

\midrule

\multirow{4}{*}{20}
& X-shape
& \klstd{36.863}{2.211}
& \klstd{7.914}{1.599}
& \klstd{7.709}{0.350}
& $\bm{\klstd{7.636}{0.290}}$
& -- \\

& GMM3
& \klstd{35.936}{2.577}
& \klstd{8.331}{0.415}
& \klstd{7.740}{0.141}
& $\bm{\klstd{7.543}{0.137}}$
& -- \\

& Funnel
& \klstd{38.551}{5.300}
& \klstd{13.766}{0.583}
& \klstd{11.773}{0.185}
& $\bm{\klstd{7.007}{0.234}}$
& -- \\

& Ring
& \klstd{42.651}{4.803}
& \klstd{9.159}{0.646}
& \klstd{8.369}{0.364}
& $\bm{\klstd{7.942}{0.140}}$
& -- \\

\bottomrule
\end{tabular}
\endgroup
\end{table}

Recall that $L$ denotes the local window size, or the number of wires acted upon by each gate tensor.
A larger $L$ allows each gate to capture more complex dependencies, but also increases the number of parameters.
Since each gate is represented by a $K^L \times K^L$ matrix, its parameter complexity scales as $O(K^{2L})$.
Importantly, this exponential dependence is on the local window size $L$, rather than the full dimension $D$.
Thus, keeping $L$ small allows \model{} to scale to high-dimensional distributions with manageable computational costs.

We study this trade-off across all non-Gaussian targets, varying $L\in\{1,\ldots,5\}$ for $D\in\{5,10\}$ and $L\in\{1,\ldots,4\}$ for $D=20$.
Table~\ref{tab:ablation_local_window} reports the results.

Across all targets, $L=1$ yields substantially worse approximation quality, highlighting the importance of modeling dependencies between variables.
Increasing $L$ to $2$ already brings substantial improvements, while larger windows provide further gains, especially in higher dimensions.
The improvement is particularly pronounced for Funnel, whose conditional-scale dependence is difficult to capture with small windows.
These results highlight the trade-off between expressiveness and computational efficiency.

\paragraph{Number of basis functions \(K\).}


\begin{table*}[t]
\centering
\caption{
Effect of basis size $K$ on forward KL divergence
(mean $\pm$ standard deviation over five runs; lower is better).
The best mean among the displayed settings in each row is highlighted in \textbf{bold}.
Complete results are provided in Table~\ref{tab:ablation_basis_size_full}.
}
\label{tab:ablation_basis_size}

\vspace{4pt}

\begingroup
\footnotesize
\setlength{\tabcolsep}{5pt}
\renewcommand{\arraystretch}{1.15}

\begin{tabular}{@{}clccccc@{}}
\toprule
$D$ & Target
& $K=1$ & $K=3$ & $K=5$ & $K=7$ & $K=9$ \\
\midrule

\multirow{4}{*}{5}
& X-shape
& \klstd{2.024}{0.010}
& \klstd{0.863}{0.023}
& \klstd{0.265}{0.007}
& \klstd{0.072}{0.004}
& \klstd{\mathbf{0.024}}{0.003} \\

& GMM3
& \klstd{2.198}{0.029}
& \klstd{0.846}{0.029}
& \klstd{0.160}{0.004}
& \klstd{0.049}{0.002}
& \klstd{\mathbf{0.020}}{0.001} \\

& Funnel
& \klstd{2.141}{0.033}
& \klstd{\mathbf{1.593}}{0.117}
& \klstd{2.838}{1.262}
& \klstd{3.488}{0.317}
& \klstd{4.247}{0.154} \\

& Ring
& \klstd{4.700}{0.017}
& \klstd{1.595}{0.028}
& \klstd{0.879}{0.018}
& \klstd{0.289}{0.004}
& \klstd{\mathbf{0.276}}{0.005} \\

\midrule

\multirow{4}{*}{10}
& X-shape
& \klstd{4.831}{0.029}
& \klstd{4.399}{0.106}
& \klstd{3.212}{0.069}
& \klstd{1.865}{0.029}
& \klstd{\mathbf{0.787}}{0.037} \\

& GMM3
& \klstd{4.984}{0.032}
& \klstd{4.486}{0.053}
& \klstd{3.146}{0.065}
& \klstd{1.772}{0.063}
& \klstd{\mathbf{0.823}}{0.078} \\

& Funnel
& \klstd{\mathbf{4.934}}{0.037}
& \klstd{5.220}{0.134}
& \klstd{7.929}{1.195}
& \klstd{9.420}{0.761}
& \klstd{11.118}{0.511} \\

& Ring
& \klstd{7.488}{0.035}
& \klstd{5.144}{0.029}
& \klstd{3.822}{0.047}
& \klstd{2.056}{0.050}
& \klstd{\mathbf{1.062}}{0.045} \\

\midrule

\multirow{4}{*}{20}
& X-shape
& \klstd{10.268}{0.088}
& \klstd{9.515}{0.209}
& \klstd{7.914}{1.599}
& \klstd{\mathbf{5.289}}{1.775}
& \klstd{5.345}{0.896} \\

& GMM3
& \klstd{10.407}{0.039}
& \klstd{10.067}{0.752}
& \klstd{8.331}{0.415}
& \klstd{6.399}{1.763}
& \klstd{\mathbf{4.801}}{1.391} \\

& Funnel
& \klstd{\mathbf{10.331}}{0.073}
& \klstd{10.671}{0.397}
& \klstd{13.766}{0.583}
& \klstd{14.322}{0.513}
& \klstd{15.645}{1.036} \\

& Ring
& \klstd{12.870}{0.052}
& \klstd{10.416}{0.190}
& \klstd{9.159}{0.646}
& \klstd{\mathbf{5.483}}{1.505}
& \klstd{5.543}{1.176} \\

\bottomrule
\end{tabular}
\endgroup
\end{table*}

Recall that $K$ denotes the number of basis functions, which determines the dimension of the local feature space.
A larger $K$ provides a richer basis for approximating the target distribution, but also increases the number of parameters in each gate tensor, which scales as $O(K^{2L})$.
We evaluate $K\in\{1,\ldots,10\}$ across all
non-Gaussian targets at $D\in\{5,10,20\}$.
Representative results for
$K\in\{1,3,5,7,9\}$ are reported in
Table~\ref{tab:ablation_basis_size},
with complete results provided in
Table~\ref{tab:ablation_basis_size_full}.

Increasing $K$ generally improves approximation accuracy on X-shape, GMM3, and Ring, although the gains vary across dimensions.
However, larger bases do not always lead to better results.
In particular, performance deteriorates sharply in several higher-dimensional settings at $K=10$, suggesting increased optimization difficulty.
Funnel exhibits a different pattern, where increasing $K$ does not consistently improve approximation accuracy.
Together with the local-window ablation, these results suggest that richer local features alone are insufficient: effective approximation also requires sufficient capacity to model dependencies between variables.

\paragraph{Ancilla size \(E\).}

\begin{table*}[t]
\centering
\caption{
Effect of the number of ancilla wires $E$ on forward KL divergence
(mean $\pm$ standard deviation over five runs; lower is better).
The best mean in each row is highlighted in \textbf{bold}.
Complete results are provided in Table~\ref{tab:ablation_ancilla_full}.
``--'' indicates untested settings.
}
\label{tab:ablation_ancilla}

\vspace{4pt}

\begingroup
\scriptsize
\setlength{\tabcolsep}{5pt}
\renewcommand{\arraystretch}{1.15}

\begin{tabular}{@{}clcccccc@{}}
\toprule
$D$ & Target
& $E=0$ & $E=2$ & $E=4$
& $E=6$ & $E=8$ & $E=10$ \\
\midrule

\multirow{4}{*}{5}
& X-shape
& \klstd{0.436}{0.162}
& \klstd{\mathbf{0.388}}{0.040}
& \klstd{0.396}{0.038}
& -- & -- & -- \\

& GMM3
& \klstd{\mathbf{0.173}}{0.023}
& \klstd{0.202}{0.033}
& \klstd{0.203}{0.014}
& -- & -- & -- \\

& Funnel
& \klstd{\mathbf{3.131}}{0.553}
& \klstd{3.763}{0.782}
& \klstd{3.193}{0.342}
& -- & -- & -- \\

& Ring
& \klstd{1.054}{0.143}
& \klstd{1.065}{0.204}
& \klstd{\mathbf{0.927}}{0.035}
& -- & -- & -- \\

\midrule

\multirow{4}{*}{10}
& X-shape
& \klstd{\mathbf{3.231}}{0.224}
& \klstd{3.253}{0.188}
& \klstd{3.339}{0.250}
& \klstd{3.277}{0.241}
& \klstd{3.385}{0.177}
& \klstd{3.366}{0.248} \\

& GMM3
& \klstd{3.405}{0.343}
& \klstd{3.216}{0.099}
& \klstd{3.250}{0.125}
& \klstd{3.340}{0.324}
& \klstd{\mathbf{3.091}}{0.166}
& \klstd{3.234}{0.136} \\

& Funnel
& \klstd{9.052}{1.045}
& \klstd{8.772}{0.581}
& \klstd{8.603}{0.808}
& \klstd{9.063}{0.322}
& \klstd{9.084}{0.486}
& \klstd{\mathbf{7.870}}{0.844} \\

& Ring
& \klstd{3.948}{0.187}
& \klstd{3.956}{0.308}
& \klstd{\mathbf{3.829}}{0.191}
& \klstd{3.966}{0.099}
& \klstd{3.953}{0.249}
& \klstd{4.052}{0.126} \\

\midrule

\multirow{4}{*}{20}
& X-shape
& \klstd{8.294}{1.607}
& \klstd{7.947}{1.502}
& \klstd{8.093}{0.189}
& \klstd{8.519}{0.607}
& \klstd{\mathbf{7.306}}{1.071}
& \klstd{8.491}{1.104} \\

& GMM3
& \klstd{8.021}{0.862}
& \klstd{7.916}{1.518}
& \klstd{8.552}{0.452}
& \klstd{8.646}{0.715}
& \klstd{\mathbf{7.648}}{1.596}
& \klstd{8.154}{1.810} \\

& Funnel
& \klstd{13.143}{1.655}
& \klstd{12.637}{1.569}
& \klstd{13.661}{0.550}
& \klstd{13.336}{1.518}
& \klstd{\mathbf{12.345}}{1.309}
& \klstd{13.843}{1.085} \\

& Ring
& \klstd{8.080}{1.967}
& \klstd{8.285}{1.318}
& \klstd{7.829}{1.846}
& \klstd{7.532}{1.904}
& \klstd{8.062}{1.856}
& \klstd{\mathbf{7.407}}{1.978} \\

\bottomrule
\end{tabular}
\endgroup
\end{table*}

Recall that $E$ denotes the number of ancilla wires, which enable \model{} to represent mixed states through purification.
Increasing $E$ expands the representational capacity of the model, but also introduces additional parameters and computational costs.
We evaluate $E\in\{0,2,4\}$ for $D=5$ and $E\in\{0,2,4,6,8,10\}$ for $D\in\{10,20\}$ across all non-Gaussian targets, with $E=0$ corresponding to the pure-state baseline.
Representative results are reported in Table~\ref{tab:ablation_ancilla}, with complete results provided in Table~\ref{tab:ablation_ancilla_full}.

Introducing ancilla wires improves approximation accuracy in several settings, demonstrating the potential benefit of mixed-state representations.
However, increasing $E$ does not consistently improve performance, and the gains vary across targets and dimensions.
Similarly, larger $E$ does not consistently reduce the variability across runs.
These results suggest that mixed-state representations can improve approximation quality, but increasing the number of ancilla wires alone does not guarantee better accuracy or more stable optimization.

\section{Related Work}

\textbf{Variational inference (VI).}
VI approximates Bayesian posteriors through KL-divergence minimization or ELBO maximization~\citep{blei2017variational,hoffman2013stochastic,ranganath2014black}.
Probabilistic circuits (PCs) provide tractable probabilistic models~\citep{shih2020probabilistic,vergari2021compositional}, with connections to tensor factorizations, sum-of-squares representations, and positive semidefinite matrix computations~\citep{loconte2025relationship,loconte2025sum,martires2025quantum}.
Score-based VI (SBVI) instead minimizes Fisher divergence using target score evaluations~\citep{hyvarinen2005estimation,yu2023semiimplicit,modi2023variational,caibatch,cai2026fisher}.
EigenVI~\citep{cai2024eigenvi} extends SBVI beyond Gaussian families through orthogonal basis expansions, reducing Fisher-divergence minimization to an eigenvalue problem.
However, its explicit tensor-product coefficients grow exponentially with dimension, and its formulation uses a single eigenvector.
QuanVI extends EigenVI to density operators and employs a compact QTN parameterization to improve scalability.

\textbf{Tensor networks (TNs).}
Tensor trains (TT/MPS) and matrix product operators (MPOs) represent high-dimensional arrays through local tensor contractions~\citep{oseledets2011tensor,orus2014practical}, with applications in density estimation~\citep{novikov2021tensor} and distributional optimization~\citep{qiu2025steps}.
Born machines construct distributions from squared amplitudes~\citep{han2018unsupervised,meiburg2025generative}, while locally purified TNs support mixed-state representations~\citep{glasser2019expressive}.
Recent theory establishes approximation hardness and polynomial-bond guarantees for MPO Born machines under suitable assumptions~\citep{li2026approximation}.
Circuit-inspired TNs have also been explored in quantum learning and state preparation~\citep{huggins2019towards,ran2020encoding}, model adaptation~\citep{chen2024quanta,li2025parameter}, and joint structure search~\citep{jointStructureSearch}.
QuanVI combines QTN purification with a Fisher-derived spectral loss and local loss approximation, enabling tractable probabilistic inference and scalable SBVI.

\section{Conclusion}
We introduced \model{}, a quantum-inspired score-based variational inference method based on density operators and quantum tensor networks.
The density-operator formulation extends EigenVI from a single eigenvector to mixed-state representations, allowing multiple eigen-directions to contribute to the variational solution.
A local approximation of the loss enables efficient estimation using low-dimensional computations.
The QTN parameterization further avoids constructing the global operator, enabling \model{} to scale to $D=100$ in our experiments.

Experiments on synthetic distributions and Bayesian posterior benchmarks demonstrate accurate approximation of Gaussian targets and competitive performance on structured non-Gaussian distributions.
Ablation studies further illustrate the trade-offs between expressiveness, computational efficiency, and optimization stability.
These results demonstrate the potential of \model{} for scalable score-based variational inference and motivate future research on adaptive QTN architectures and more general dependency structures.

\section{Limitations}

\model{} uses a local approximation of the loss, which may introduce errors for distributions with long-range dependencies.
Its expressiveness depends on the QTN architecture, while increasing model capacity raises computational costs without guaranteeing better accuracy.
Training can also be slower than simpler variational baselines.
Finally, broader Bayesian benchmarks with matched computational budgets are needed to assess its practical advantages.



\section*{Acknowledgments}
YC, CL, ZS, and QZ were supported by JSPS KAKENHI Grant Number JP24K03005.
ZT was supported by JSPS KAKENHI Grant Number JP26K21321 and the RIKEN Incentive Research Project.
QZ was additionally supported by JSPS KAKENHI Grant Number JP23K28109 and the RIKEN TRIP initiative (RIKEN Quantum).
YC, CL, and ZS also received support from the JHPC-quantum Test User Program.

{\small
\bibliographystyle{unsrtnat}
\bibliography{Ref}
}
\clearpage

\appendix
\section{Preliminary}
\subsection{Derivation of \(M\)}
\label{app:eigenvi_details}

We derive the quadratic form in Eq.~\eqref{eq:cai_eigen}. Let
\(a_\theta(x) := \langle \theta,\Phi(x)\rangle\), so that
\(q_\theta=a_\theta^2\) and
\(\nabla_x \log q_\theta = 2\nabla_x a_\theta/a_\theta\).
Define the derivative matrix \(\dot{\Phi}(x)\in\mathbb{R}^{D\times K^D}\) by
\begin{equation}
    \dot{\Phi}(x)=
    \begin{pmatrix}
        \dot{\phi}(x_1)\otimes
        \left[\bigotimes_{d=2}^D \phi(x_d)^\top\right]
        \\
        \vdots
        \\
        \left[\bigotimes_{d=1}^{\ell-1}\phi(x_d)^\top\right]
        \otimes
        \dot{\phi}(x_\ell)
        \otimes
        \left[\bigotimes_{d=\ell+1}^D\phi(x_d)^\top\right]
        \\
        \vdots
        \\
        \left[\bigotimes_{d=1}^{D-1}\phi(x_d)^\top\right]
        \otimes
        \dot{\phi}(x_D)
    \end{pmatrix},
    \label{eq:app_dot_phi}
\end{equation}
where
\[
    \dot{\phi}(x_d)
    :=
    \left[
        \frac{d\phi_1(x_d)}{dx_d},
        \frac{d\phi_2(x_d)}{dx_d},
        \ldots,
        \frac{d\phi_K(x_d)}{dx_d}
    \right]
    \in \mathbb{R}^{1\times K}.
\]
Thus, \(\nabla_x a_\theta(x)=\dot{\Phi}(x)\theta\). Let
\(s(x):=\nabla_x\log p(x)\). Then
\[
\begin{aligned}
    q_\theta(x)
    \left\|
        \nabla_x\log p(x)
        -
        \nabla_x\log q_\theta(x)
    \right\|_F^2
    &=
    a_\theta(x)^2
    \left\|
        s(x)
        -
        2\frac{\dot{\Phi}(x)\theta}{a_\theta(x)}
    \right\|_F^2 \\
    &=
    \left\|
        a_\theta(x)s(x)
        -
        2\dot{\Phi}(x)\theta
    \right\|_F^2 \\
    &=
    \left\|
        \left(
            s(x)\Phi(x)^\top
            -
            2\dot{\Phi}(x)
        \right)\theta
    \right\|_F^2 .
\end{aligned}
\]
Integrating over \(x\) gives
\(\int q_\theta\|s-\nabla_x\log q_\theta\|^2dx=\theta^\top M\theta\), with
\(M\) defined in Eq.~\eqref{eq:def_M}; that is, the objective in
Eq.~\eqref{eq:fisher} equals \(\frac12\theta^\top M\theta\). Since \(M\) is an
integral of Gram matrices, it is positive semidefinite.

\section{Detailed Complexity Analysis}
\label{app:complexity}

Let \(D\) be the number of variables, \(E\) the number of ancilla wires,
\(K\) the wire dimension, \(L\) the local window size, and \(N\) the number
of generated samples. The implementation uses
\[
J = D + E - L + 1
\]
gates, each containing \(K^{2L}\) entries. Let \(B\) denote the total
proposal-sample budget used to estimate all local operators in one loss
evaluation. In the implementation, this budget is set to
\(\max(B,100D)\) and divided across the \(D\) local operators.

\paragraph{Memory.}
The trainable gates require \(O(JK^{2L})\) memory. The stored local operators
require \(O(DK^6)\) memory, since each acts on at most three variables.
The full density operator would require \(O(K^{2D})\) memory, which the
tensor-network representation avoids.

\paragraph{Training.}
At each optimization step, the code estimates \(D\) local operators using a
total of at most \(\max(B,100D)\) proposal samples, resampling the operators
at each step. The local Gram-matrix construction costs
\[
O\!\left(\max(B,100D)(DK+K^6)\right),
\]
excluding the cost of evaluating the target score. This corresponds to
\(\max(B,100D)\) full \(D\)-dimensional score evaluations per step.

The loss is a sum of \(D\) local contractions. Under the contraction order
used by the implementation, their cost is bounded by
\(O(DJ K^{O(L)})\) per step. The SGDG update applies a Cayley step to each
\(K^L\times K^L\) gate, costing \(O(JK^{3L})\). Thus, for \(T\) optimization
steps, the total cost is
\[
O\!\left(
T\left[
\max(B,100D)(DK+K^6)
+ DJ K^{O(L)}
+ J K^{3L}
\right]
\right),
\]
in addition to the target-score evaluation cost. For fixed \(K\) and \(L\),
the tensor-network and local-operator costs grow polynomially with \(D\)
and \(E\).

\paragraph{Inference.}
A density query at one point requires a tensor-network contraction costing
\(O(JK^{3L})\) under the contraction scheme used here. Autoregressive
sampling evaluates a conditional density on an \(n_{\mathrm{grid}}\)-point
grid for each sample and coordinate. Drawing \(N\) samples therefore costs
\[
O\!\left(ND n_{\mathrm{grid}} J K^{3L}\right),
\]
plus the cost of interpolation on the grid. The grid range is \([-5,5]\);
the scripts use 100 grid points for synthetic runs and 500 for PosteriorDB
runs.

\section{Experiments}

\subsection{Computational resources}
The synthetic benchmark experiments were run on CPU nodes with AMD EPYC 9654
processors and 1.5 TiB memory. The posterior benchmark experiments were run on
CPU nodes with Intel Xeon Silver 4316 processors. The ablation studies were run
on 112-core CPU nodes with 128 GB memory. No GPUs were used.

\subsection{Synthetic experiments}
\begin{figure*}[t]
    \centering
    \begin{subfigure}{0.19\textwidth}
        \centering
        \includegraphics[width=\linewidth]{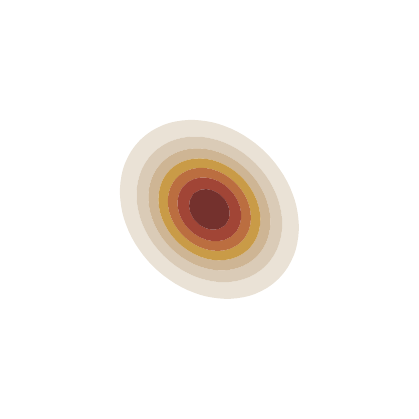}
        \caption{Gaussian}
    \end{subfigure}
    \hfill
    \begin{subfigure}{0.19\textwidth}
        \centering
        \includegraphics[width=\linewidth]
        {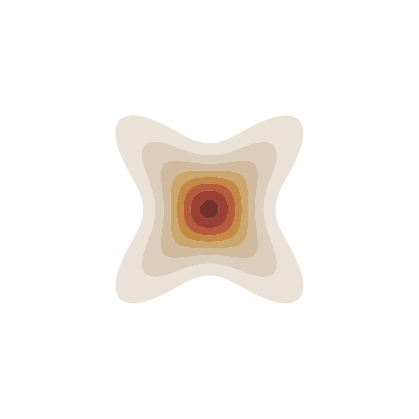}
        \caption{X-shape}
    \end{subfigure}
    \hfill
    \begin{subfigure}{0.19\textwidth}
        \centering
        \includegraphics[width=\linewidth]
        {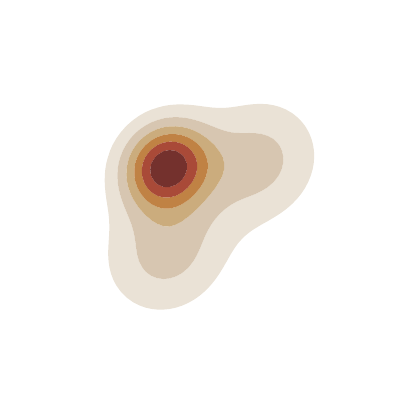}
        \caption{GMM3}
    \end{subfigure}
    \hfill
    \begin{subfigure}{0.19\textwidth}
        \centering
        \includegraphics[width=\linewidth]
        {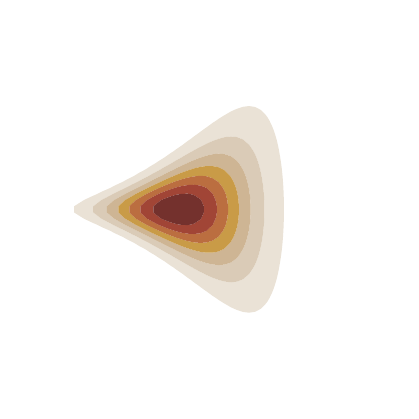}
        \caption{Funnel}
    \end{subfigure}
    \hfill
    \begin{subfigure}{0.19\textwidth}
        \centering
        \includegraphics[width=\linewidth]
        {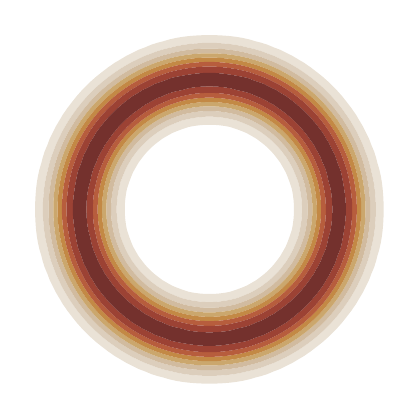}
        \caption{Ring}
    \end{subfigure}
    \caption{Two-dimensional marginal distributions over the first two variables of the synthetic targets used in our experiments. Darker colors indicate higher probability density within each distribution.}
    \label{fig:synthetic-targets}
\end{figure*}

\subsubsection{Two-dimensional base targets}
\label{app:synthetic_base_targets}

We specify the four two-dimensional non-Gaussian base targets shown in
Fig.~\ref{fig:synthetic-targets}.
Let \(u=(x_1,x_2)^\top\in\mathbb{R}^2\), and write \(p_0(u)\) for the
corresponding base density.
Throughout, \(\mathcal{N}(\,\cdot\,;\mu,\Sigma)\) denotes a Gaussian density
with mean \(\mu\) and covariance \(\Sigma\); for a univariate Gaussian,
the second parameter denotes the variance.

\paragraph{GMM3.}
The three-component Gaussian mixture is
\[
p_0(u)
=
0.4\,\mathcal{N}\!\left(u;\mu_1,\tfrac{1}{2}\Sigma_0\right)
+0.3\,\mathcal{N}(u;\mu_2,\Sigma_0)
+0.3\,\mathcal{N}(u;\mu_3,\Sigma_0),
\]
where
\[
\mu_1=\begin{bmatrix}-1\\1\end{bmatrix},
\qquad
\mu_2=\begin{bmatrix}1.1\\1.1\end{bmatrix},
\qquad
\mu_3=\begin{bmatrix}-1\\-1\end{bmatrix},
\qquad
\Sigma_0=\begin{bmatrix}1&0.1\\0.1&1\end{bmatrix}.
\]

\paragraph{X-shape.}
The X-shaped target is an equally weighted mixture of two zero-mean
Gaussians:
\[
p_0(u)
=
\frac{1}{2}\mathcal{N}(u;0,\Sigma_+)
+
\frac{1}{2}\mathcal{N}(u;0,\Sigma_-),
\]
with
\[
\Sigma_+
=
\begin{bmatrix}
1.65&1.35\\
1.35&1.65
\end{bmatrix},
\qquad
\Sigma_-
=
\begin{bmatrix}
1.65&-1.35\\
-1.35&1.65
\end{bmatrix}.
\]
These covariance matrices are obtained by rotating
\(\operatorname{diag}(3,0.3)\) by \(+\pi/4\) and \(-\pi/4\), respectively.

\paragraph{Ring.}
The ring-shaped target has density
\[
p_0(u)
=
\frac{1}{Z_{\mathrm{ring}}}
\exp\!\left(
-\frac{(\|u\|_2-r_0)^2}{2\sigma_{\mathrm{ring}}^2}
\right),
\qquad
r_0=3,\quad \sigma_{\mathrm{ring}}=0.5,
\]
where the normalizing constant is
\[
Z_{\mathrm{ring}}
=
2\pi\int_0^\infty
r\exp\!\left(
-\frac{(r-r_0)^2}{2\sigma_{\mathrm{ring}}^2}
\right)\,dr.
\]
The density is rotationally symmetric and concentrates around the
circle of radius \(r_0\).

\paragraph{Funnel.}
The funnel target is defined by
\[
x_1\sim\mathcal{N}(0,\sigma_{\mathrm{f}}^2),
\qquad
x_2\mid x_1\sim\mathcal{N}(0,e^{x_1}),
\qquad
\sigma_{\mathrm{f}}=1.2.
\]
Equivalently, its joint density is
\[
p_0(u)
=
\mathcal{N}(x_1;0,\sigma_{\mathrm{f}}^2)\,
\mathcal{N}(x_2;0,e^{x_1}).
\]
The conditional standard deviation of \(x_2\) is \(e^{x_1/2}\),
producing a narrow neck for negative \(x_1\) and an increasingly
dispersed region for positive \(x_1\).

\subsubsection{Construction of High-dimensional Synthetic Targets}
\label{app:high-dim-transform}

\begin{figure}[t]
    \centering
    \includegraphics[width=0.85\linewidth]{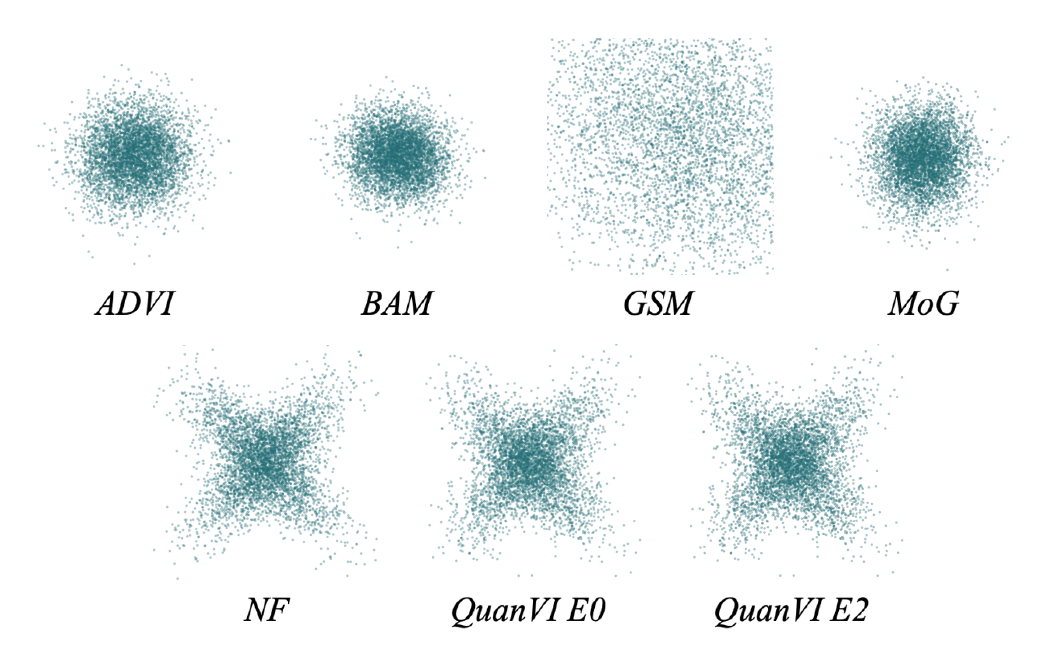}
    \caption{Sample visualization on the first two dimensions of the X-shape target with $D=20$, comparing all baselines with our method.}
    \label{fig:xshape_d20}
    \vspace{-1em}
\end{figure}

We describe the construction of the high-dimensional synthetic targets used in
the experiments. For the non-Gaussian targets, namely \textbf{GMM3},
\textbf{X-shape}, \textbf{Ring}, and \textbf{Funnel}, we start from a
two-dimensional base target and augment it with a nonlinear Markov chain. The
\textbf{Gaussian} target is constructed directly in \(D\) dimensions as a
locally correlated Gaussian distribution, as described at the end of this
section.

Let \(p_0(x_1,x_2)\) denote one of the base two-dimensional non-Gaussian target
densities in Fig.~\ref{fig:synthetic-targets}. For ambient dimension
\(D\ge 2\), define \(N=D-2\) auxiliary variables
\(z_1,\ldots,z_N\), with \(z_0=x_2\).
The augmented target is
\[
p(x_1,x_2,z_1,\ldots,z_N)
=
p_0(x_1,x_2)
\prod_{i=1}^{N}
\mathcal{N}\!\left(
z_i;\,h_i(z_{i-1}),\,\sigma_{\mathrm{aug}}^2
\right).
\]
Hence the dependency structure is a single chain
\(x_2\to z_1\to z_2\to\cdots\to z_N\), i.e., in coordinate indices
\(2\to3\to\cdots\to D\).

Each \(h_i:\mathbb{R}\to\mathbb{R}\) is selected deterministically from
a fixed smooth function bank:
\[
\begin{aligned}
h(x)\in\Big\{
&\pm 1.5\sin(x),\,
\pm 1.5\cos(x),\,
\pm 1.5\sin(2x),\,
\pm 1.5\cos(2x),\\
&\pm 1.5\big(\sigma(4x)-1/2\big),\,
\pm 1.5\tanh(2x),\,
\pm 1.5\tanh(4x),\\
&\pm 1.5\sin(x)\tanh(x),\,
\pm 1.5\cos(x)\tanh(x),\,
\pm 1.5\,\frac{x}{1+x^2}
\Big\},
\end{aligned}
\]
where \(\sigma(\cdot)\) is the logistic sigmoid.
In implementation, this bank is expanded deterministically by repetition,
and we use the first \(N=D-2\) transforms as \(h_1,\ldots,h_N\).
Therefore, when \(D\) increases, the transform pattern continues in the
same fixed order.

Sampling is sequential:
\[
(x_1,x_2)\sim p_0,\qquad
z_i = h_i(z_{i-1}) + \sigma_{\mathrm{aug}}\epsilon_i,\quad
\epsilon_i\sim\mathcal N(0,1),\quad i=1,\ldots,N.
\]
Thus the original two-dimensional geometry is preserved in the marginal
\((x_1,x_2)\), while higher coordinates introduce nonlinear local
dependencies along the chain.

The log-density is available in closed form:
\[
\log p(x_1,x_2,z_1,\ldots,z_N)
=
\log p_0(x_1,x_2)
-\frac{N}{2}\log(2\pi\sigma_{\mathrm{aug}}^2)
-\frac{1}{2\sigma_{\mathrm{aug}}^2}
\sum_{i=1}^{N}\!\left(z_i-h_i(z_{i-1})\right)^2.
\]
This expression is used for exact density evaluation. The score is also
analytic from the Markov factorization. Let
\(s_0(x_1,x_2)=\nabla_{(x_1,x_2)}\log p_0(x_1,x_2)\). Then
\[
\begin{aligned}
\nabla_{x_1}\log p
&=
[s_0(x_1,x_2)]_1,\\
\nabla_{x_2}\log p
&=
[s_0(x_1,x_2)]_2
+\frac{h_1'(x_2)}{\sigma_{\mathrm{aug}}^2}
\left(z_1-h_1(x_2)\right),
\end{aligned}
\]
and for \(i=1,\ldots,N\),

\[
\nabla_{z_i}\log p
=
\frac{h_i(z_{i-1})-z_i}{\sigma_{\mathrm{aug}}^2}
+\mathbf{1}_{\{i<N\}}
\frac{h_{i+1}'(z_i)}{\sigma_{\mathrm{aug}}^2}
\left(z_{i+1}-h_{i+1}(z_i)\right).
\]
The first term is the parent contribution
\(\log p(z_i\mid z_{i-1})\), and the second term, when \(i<N\), is the child
contribution from \(\log p(z_{i+1}\mid z_i)\).

Overall, this construction yields high-dimensional targets that retain
the non-Gaussian two-dimensional structure in \((x_1,x_2)\), while
adding controllable nonlinear local dependencies through a single
path-structured chain. Exact sampling, exact log-density, and exact
score are all available, and both sampling and score evaluation scale
linearly with \(D\).

\paragraph{Gaussian target.}
The Gaussian target is constructed directly in \(D\) dimensions rather than by
augmenting a two-dimensional base distribution. Specifically, we use
\[
    p(x)=\mathcal{N}(x;0,\Sigma),
    \qquad
    \Sigma=\Lambda^{-1},
\]
where \(\Lambda\in\mathbb{R}^{D\times D}\) is a tridiagonal precision matrix
with diagonal entries \(1\) and neighboring off-diagonal entries \(0.2\):
\[
    \Lambda_{ii}=1,\qquad
    \Lambda_{i,i+1}=\Lambda_{i+1,i}=0.2.
\]
All other entries are zero. This construction gives a locally correlated
Gaussian distribution whose dependency structure follows a path graph, matching
the Markov dependency assumption used in the main text. The log-density and
score are available in closed form:
\[
    \log p(x)
    =
    -\frac{1}{2}x^\top \Lambda x
    + C,
    \qquad
    \nabla_x\log p(x)
    =
    -\Lambda x,
\]
where \(C\) is the normalizing constant.

\subsubsection{Hyperparameter settings for synthetic experiments}
\label{syn_para}

We summarize the hyperparameter settings used for the synthetic experiments.
The target distributions are
\texttt{xshape}, \texttt{ring}, \texttt{funnel}, \texttt{gmm3}, and
\texttt{gaussian}. Unless otherwise stated, the ambient dimensions are
\[
    D \in \{5,10,20,100\}.
\]
Each configuration is repeated with five independent random seeds, and the
evaluation follows Section~\ref{sec:exp_synthetic}. Following EigenVI, both
EigenVI and QuanVI use normalized Hermite functions as the basis \(\phi\).

\paragraph{GSM.}
We run Gaussian score matching (GSM) with the Gaussian variational family.
For each target and dimension, we use \(2000\) iterations. No learning rate or
mini-batch size is required for this baseline.

\paragraph{BaM.}
We run Batch-and-Match VI (BaM) with the Gaussian variational family. The
optimization budget is \(2000\) iterations. The batch size is set to
\[
    B_{\mathrm{BaM}} = \max(1,D-1),
\]
which is the largest value satisfying the implementation constraint
\(B_{\mathrm{BaM}} < D\).

\paragraph{ADVI.}
We run ADVI with the same iteration budget as BaM, namely \(2000\) iterations.
The batch size is also set to
\[
    B_{\mathrm{ADVI}} = \max(1,D-1).
\]
Other optimizer parameters are kept at their default implementation values.

\paragraph{NF.}
We run the normalizing-flow baseline using the same synthetic targets and
dimensions. The optimization budget is \(2000\) iterations. The number of
Monte Carlo samples used by the NF baseline is set as
\[
    N_{\mathrm{NF}}
    =
    \mathrm{clip}(256D,\;512,\;8192).
\]
Thus, for \(D\in\{5,10,20,100\}\), this gives
\[
    N_{\mathrm{NF}} \in \{1280, 2560, 5120, 8192\}.
\]
Other architecture and optimizer parameters are kept at the default values of
the NF implementation.

\paragraph{EigenVI.}
We use \(K=5\) basis functions, the same as QuanVI. The importance proposal
is the product uniform distribution \(\mathrm{Unif}([-5,5])^{\otimes D}\),
and the number of importance samples is
\(N_{\mathrm{EigenVI}}=\mathrm{clip}(2000D,\,10^4,\,2\times10^5)\).
For \(D>5\), the global \(K^D\times K^D\) matrix \(M\) cannot be stored in
the available memory; for example, \(K^D\approx 9.8\times10^{6}\) at
\(D=10\). The corresponding entries are reported as OOM in
Table~\ref{tab:synthetic_dimension_scaling}.

\paragraph{MoG.}
We run mixture-of-Gaussians VI with isotropic covariance components. The default
learning rate is \(10^{-2}\), the default number of iterations is \(2000\), the
default number of mixture components is \(5\), and the default gradient batch
size is \(10\). For more challenging targets, we use the following
target-specific settings:
\[
K_{\mathrm{mix}} =
\begin{cases}
10, & \texttt{gmm3} \text{ and } \texttt{ring},\\
20, & \texttt{funnel},\\
5,  & \text{otherwise},
\end{cases}
\]
and
\[
B_{\mathrm{grad}} =
\begin{cases}
15, & \texttt{gmm3} \text{ and } \texttt{ring},\\
20, & \texttt{funnel},\\
10, & \text{otherwise}.
\end{cases}
\]
For the \texttt{funnel} target, we also reduce the learning rate to \(10^{-3}\)
and increase the number of iterations to \(3500\). The KL estimation batch size
is \(B_{\mathrm{KL}}=1000\). We set
\texttt{vgmm\_scale\_cov}\(=5\) and \texttt{vgmm\_sample\_boule}\(=5\).

\paragraph{QuanVI.}
For QuanVI, we use the same synthetic targets and dimensions as above. We run
both the pure-state and mixed-state variants by setting
\[
    E \in \{0,2\},
\]
where \(E=0\) corresponds to a pure-state ansatz and \(E=2\) introduces ancilla
degrees of freedom for mixed-state representations. At each optimization step,
the total number of samples used to estimate all \(D\) local operators scales
linearly with dimension:
\[
    B_{\mathrm{QuanVI}} = 500D.
\]
This total budget is divided evenly across the \(D\) local operators, giving
about 500 samples per operator at each step. For example,
\(B_{\mathrm{QuanVI}}=10^4\) when \(D=20\), and
\(B_{\mathrm{QuanVI}}=5\times 10^4\) when \(D=100\). We use \(K=5\) basis
functions and local window size \(L=2\), and train for 2000 optimization steps
without target-specific tuning. All other optimizer parameters are kept at the
default values of the implementation unless explicitly stated otherwise.

\subsubsection{Result of NF}
\label{app:nf}

We report the normalizing-flow baseline results in Table~\ref{tab:nf_synthetic}
for completeness. Although NF achieves low KL divergence on some low-dimensional
targets, its results are highly unstable across different targets and dimensions.
In several high-dimensional or strongly non-Gaussian settings, the KL divergence
becomes extremely large, indicating numerical or optimization instability. This
sensitivity makes NF difficult to use as a reliable main-table baseline in our
experimental setting. Therefore, we omit NF from the main comparison table and
include the detailed results here in the appendix.

\begin{table}[t]
\centering
\caption{Normalizing flow results on synthetic targets over five runs. Values report the median forward KL divergence, with the number of divergent runs in parentheses.}
\label{tab:nf_synthetic}
\scriptsize
\setlength{\tabcolsep}{6pt}
\begin{tabular}{lccccc}
\toprule
\(D\) & Gaussian & X-shape & GMM3 & Funnel & Ring \\
\midrule
5
& 0.0109 (0/5)
& 0.0319 (0/5)
& 0.0297 (0/5)
& 0.0778 (2/5)
& 0.2682 (0/5) \\
10
& 0.0247 (0/5)
& \(4.36\times10^{17}\) (5/5)
& 0.6085 (1/5)
& \(1.05\times10^{15}\) (4/5)
& \(6.25\times10^{28}\) (3/5) \\
20
& 0.0224 (0/5)
& 184.8571 (1/5)
& \(5.06\times10^{13}\) (3/5)
& 1.7613 (0/5)
& 3.7114 (1/5) \\
100
& 0.0774 (1/5)
& 37.5930 (1/5)
& 404.6059 (2/5)
& \(3.38\times10^{23}\) (4/5)
& 44.0952 (0/5) \\
\bottomrule
\end{tabular}
\end{table}

\subsection{Posterior experiments}
\label{app:pos}

\begin{figure}[t]
    \centering
    \includegraphics[width=\linewidth]{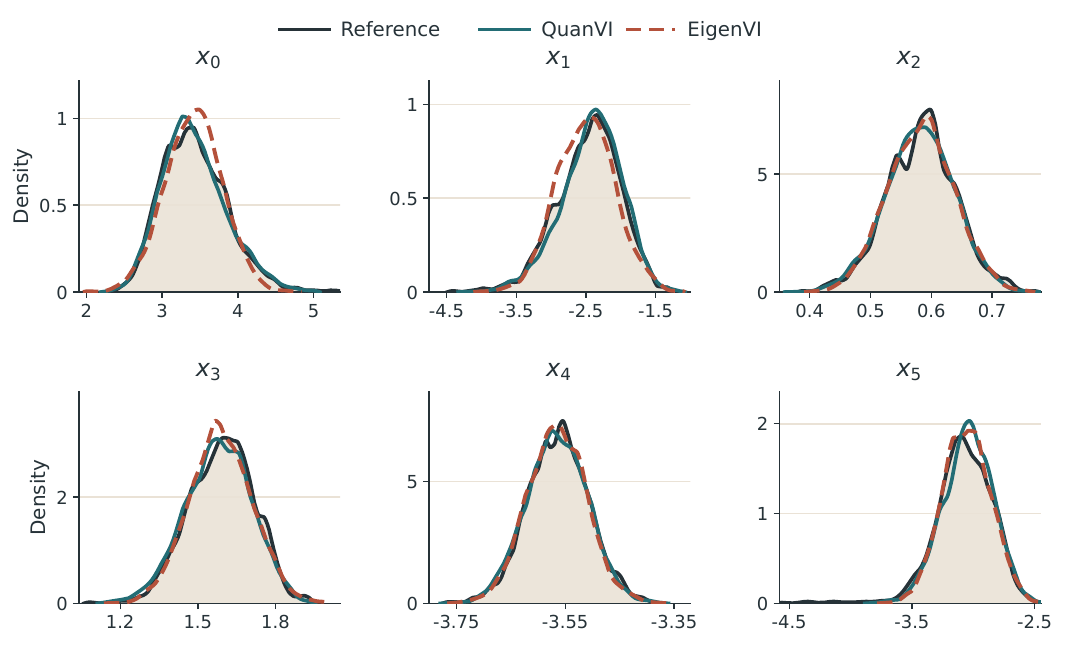}
    \caption{One-dimensional marginals of the six coordinates of
    \texttt{hmm\_bball\_0}, comparing QuanVI and EigenVI with
    reference posterior samples.}
    \label{fig:hmm_bball_0}
\end{figure}

\subsubsection{Standardization}
\label{app:standardization}

For the \textsc{posteriordb} experiments, we apply a two-stage
reparameterization before variational training. First, the constrained Stan
parameters $\boldsymbol{\theta}$ are mapped to unconstrained coordinates
$\mathbf{x}\in\mathbb{R}^D$ using Stan's bijective transformation. The target
log-density and score in this space are evaluated through \textsc{BridgeStan}
with the Jacobian adjustment enabled.

We then standardize $\mathbf{x}$ using a full-covariance Gaussian
preconditioner $\mathcal{N}(\widehat{\mathbf{m}},
\widehat{\boldsymbol{\Sigma}})$. Let
$\widehat{\boldsymbol{\Sigma}}=\mathbf{L}\mathbf{L}^{\top}$. We define
\[
\mathbf{y}
=
\mathbf{L}^{-1}(\mathbf{x}-\widehat{\mathbf{m}}),
\qquad
\mathbf{x}
=
\widehat{\mathbf{m}}+\mathbf{L}\mathbf{y}.
\]
The corresponding target score in standardized coordinates is

\[
\nabla_{\mathbf{y}}\log \tilde{p}(\mathbf{y})
=
\mathbf{L}^{\top}
\nabla_{\mathbf{x}}\log p(\mathbf{x}),
\qquad
\mathbf{x}=\widehat{\mathbf{m}}+\mathbf{L}\mathbf{y}.
\]

The preconditioner $(\widehat{\mathbf{m}},
\widehat{\boldsymbol{\Sigma}})$ is obtained before QuanVI training using GSM
for $10^4$ iterations with batch size $16$; BaM is used for \texttt{hmm\_bball\_0}, where
GSM is numerically unstable. Reference posterior draws are not used to construct
the preconditioner. The same preconditioner is shared by all methods that use
standardization.

QuanVI is trained directly in the standardized coordinates $\mathbf{y}$:
basis functions are evaluated in this space and collocation points are
sampled from $[-5,5]^D$. For evaluation, the learned score is transformed back
to the unconstrained coordinates as
$\nabla_{\mathbf{x}}\log q(\mathbf{x})
=
\mathbf{L}^{-\top}
\nabla_{\mathbf{y}}\log \tilde{q}(\mathbf{y})$.
All reported Fisher divergences are computed in this common unconstrained
$\mathbf{x}$-space. Synthetic targets are already constructed on a comparable
scale and therefore use the identity transformation.

\subsubsection{Hyperparameter settings for posterior benchmark experiments}
\label{app:posterior_hparams}

We summarize the hyperparameter settings used for the posterior benchmark
experiments. We evaluate three PosteriorDB targets, denoted by the short names
\texttt{hmm}, \texttt{hmm\_bball\_0}, and \texttt{gpregr}. These short names are
mapped to the corresponding PosteriorDB models as follows:
\[
\begin{array}{ll}
\toprule
\text{Short name} & \text{PosteriorDB model} \\
\midrule
\texttt{hmm} & \texttt{hmm\_example-hmm\_example} \\
\texttt{hmm\_bball\_0} & \texttt{bball\_drive\_event\_0-hmm\_drive\_0} \\
\texttt{gpregr} & \texttt{gp\_pois\_regr-gp\_regr} \\
\bottomrule
\end{array}
\]
Each configuration is repeated with four independent random seeds, and we report the
mean and standard deviation across runs. Unless otherwise stated, the same
hyperparameters are used for all three targets. Different from the synthetic
experiments, several posterior baselines use Gaussian initialization or
standardization obtained from GSM or BaM fits.

\paragraph{GSM.}
We run Gaussian score matching with the Gaussian variational family. GSM is used
both as a baseline and as an initialization method for other posterior baselines.
We use \(10000\) iterations and batch size \(16\).

\paragraph{BaM.}
We run Batch-and-Match VI with the Gaussian variational family. We use \(1000\)
iterations and batch size \(16\). BaM is initialized from the GSM-fitted mean and
covariance with \texttt{init\_cov\_scale}\(=1.0\) and
\texttt{init\_cov\_jitter}\(=10^{-6}\).

\paragraph{ADVI.}
We run ADVI for \(1000\) iterations with batch size \(16\) and learning rate
\(10^{-4}\). ADVI is initialized from the GSM-fitted mean and covariance with
\texttt{init\_cov\_scale}\(=1.0\) and
\texttt{init\_cov\_jitter}\(=10^{-6}\).

\paragraph{NF.}
We run the normalizing-flow baseline without using Gaussian initialization files.
The flow uses hidden dimension \(32\), \(8\) layers, \(1000\) optimization
iterations, \(16\) Monte Carlo samples, learning rate \(10^{-3}\), and
\(\beta=1.0\).

\paragraph{EigenVI.}
Since the size of the global eigenvalue problem grows as \(K^D\), we use a
dimension-dependent basis size: \(K=4\) for \texttt{gpregr} and \texttt{hmm}
(\(D\le5\)), and \(K=2\) for \texttt{hmm\_bball\_0} (\(D=6\)). We use
\(60000\) importance samples, and the target is standardized as described in
Appendix~\ref{app:standardization}.

\paragraph{MoG.}
We run mixture-of-Gaussians VI with isotropic covariance components and natural
gradient covariance updates. We use
\[
\texttt{mode}=\texttt{iso},\qquad
\texttt{variance\_update}=\texttt{ngd},\qquad
\eta=10^{-4}.
\]
The number of iterations is \(10000\), the number of mixture components is
\(50\), the KL batch size is \(B_{\mathrm{KL}}=1000\), and the gradient batch
size is \(B_{\mathrm{grad}}=20\). We set
\texttt{vgmm\_sample\_boule}\(=1\) and
\texttt{vgmm\_scale\_cov}\(=1\). MoG is initialized from GSM-fitted Gaussian
statistics whenever available, with
\texttt{init\_cov\_scale}\(=1.0\),
\texttt{init\_mean\_scale}\(=0.25\),
\texttt{init\_component\_cov\_scale}\(=0.5\), and
\texttt{init\_cov\_jitter}\(=10^{-6}\). If these files are unavailable, MoG
falls back to random initialization.

\paragraph{QuanVI.}
QuanVI uses the same standardization as EigenVI, with
\(E\in\{0,2\}\), \(K=5\), and \(B=5000\) training samples. The model is
trained for \(1000\) steps with learning rate \(10^{-1}\), momentum \(0.9\),
and complex double precision. Samples are drawn by inverse-CDF sampling on a
500-point grid.

\subsection{Ablation study}
\label{app:ablation_details}

All ablations use the non-Gaussian targets of
Appendix~\ref{app:high-dim-transform} at \(D\in\{5,10,20\}\), five random
seeds, and forward KL estimated from \(10^4\) target samples. Each ablation
varies one hyperparameter while fixing the others at \(L=2\), \(K=5\), and
\(E=0\).
The varied ranges are given in
Section~\ref{sec:exp_ablation}. To cover a broad range of settings, every
configuration is trained for 1000 steps instead of 2000, so some
configurations may not have fully converged.
While representative results are presented in the main text,
the complete results for the basis size \(K\) and ancilla size \(E\)
are provided in Tables~\ref{tab:ablation_basis_size_full}
and~\ref{tab:ablation_ancilla_full}, respectively.

\begin{table*}[t]
\centering

\caption{
Complete ablation results for basis size
$K\in\{1,\ldots,10\}$ on forward KL divergence
(mean $\pm$ std over five runs; lower is better).
Bold indicates the lowest mean in each row.
}

\label{tab:ablation_basis_size_full}
\footnotesize
\setlength{\tabcolsep}{3pt}
\renewcommand{\arraystretch}{1.15}
\resizebox{\textwidth}{!}{%
\begin{tabular}{@{}cl*{10}{c}@{}}
\toprule
\(D\) & Target
& \(K=1\) & \(K=2\) & \(K=3\) & \(K=4\) & \(K=5\)
& \(K=6\) & \(K=7\) & \(K=8\) & \(K=9\) & \(K=10\) \\
\midrule

\multirow{4}{*}{5}
& X-shape
& \klstd{2.024}{0.010}
& \klstd{1.282}{0.022}
& \klstd{0.863}{0.023}
& \klstd{0.433}{0.007}
& \klstd{0.265}{0.007}
& \klstd{0.157}{0.005}
& \klstd{0.072}{0.004}
& \klstd{0.033}{0.003}
& \klstd{0.024}{0.003}
& \klstd{\mathbf{0.012}}{0.001} \\

& GMM3
& \klstd{2.198}{0.029}
& \klstd{1.338}{0.015}
& \klstd{0.846}{0.029}
& \klstd{0.363}{0.011}
& \klstd{0.160}{0.004}
& \klstd{0.145}{0.006}
& \klstd{0.049}{0.002}
& \klstd{0.029}{0.004}
& \klstd{0.020}{0.001}
& \klstd{\mathbf{0.008}}{0.000} \\

& Funnel
& \klstd{2.141}{0.033}
& \klstd{2.059}{0.282}
& \klstd{\mathbf{1.593}}{0.117}
& \klstd{3.937}{1.857}
& \klstd{2.838}{1.262}
& \klstd{2.876}{0.181}
& \klstd{3.488}{0.317}
& \klstd{3.927}{0.074}
& \klstd{4.247}{0.154}
& \klstd{4.629}{0.348} \\

& Ring
& \klstd{4.700}{0.017}
& \klstd{2.441}{0.028}
& \klstd{1.595}{0.028}
& \klstd{1.585}{0.032}
& \klstd{0.879}{0.018}
& \klstd{0.447}{0.013}
& \klstd{0.289}{0.004}
& \klstd{1.025}{0.025}
& \klstd{0.276}{0.005}
& \klstd{\mathbf{0.108}}{0.004} \\

\midrule

\multirow{4}{*}{10}
& X-shape
& \klstd{4.831}{0.029}
& \klstd{4.453}{0.024}
& \klstd{4.399}{0.106}
& \klstd{3.962}{0.153}
& \klstd{3.212}{0.069}
& \klstd{2.923}{0.092}
& \klstd{1.865}{0.029}
& \klstd{0.948}{0.036}
& \klstd{\mathbf{0.787}}{0.037}
& \klstd{14.654}{1.949} \\

& GMM3
& \klstd{4.984}{0.032}
& \klstd{4.486}{0.057}
& \klstd{4.486}{0.053}
& \klstd{3.994}{0.077}
& \klstd{3.146}{0.065}
& \klstd{2.957}{0.039}
& \klstd{1.772}{0.063}
& \klstd{1.005}{0.042}
& \klstd{\mathbf{0.823}}{0.078}
& \klstd{13.896}{0.996} \\

& Funnel
& \klstd{\mathbf{4.934}}{0.037}
& \klstd{5.106}{0.197}
& \klstd{5.220}{0.134}
& \klstd{7.851}{1.292}
& \klstd{7.929}{1.195}
& \klstd{8.453}{0.844}
& \klstd{9.420}{0.761}
& \klstd{8.129}{1.184}
& \klstd{11.118}{0.511}
& \klstd{12.922}{0.716} \\

& Ring
& \klstd{7.488}{0.035}
& \klstd{5.577}{0.095}
& \klstd{5.144}{0.029}
& \klstd{5.125}{0.102}
& \klstd{3.822}{0.047}
& \klstd{3.241}{0.131}
& \klstd{2.056}{0.050}
& \klstd{1.937}{0.047}
& \klstd{\mathbf{1.062}}{0.045}
& \klstd{15.105}{1.830} \\

\midrule

\multirow{4}{*}{20}
& X-shape
& \klstd{10.268}{0.088}
& \klstd{6.410}{0.106}
& \klstd{9.515}{0.209}
& \klstd{8.481}{1.162}
& \klstd{7.914}{1.599}
& \klstd{7.996}{0.677}
& \klstd{\mathbf{5.289}}{1.775}
& \klstd{5.751}{1.048}
& \klstd{5.345}{0.896}
& \klstd{25.636}{4.765} \\

& GMM3
& \klstd{10.407}{0.039}
& \klstd{6.360}{0.054}
& \klstd{10.067}{0.752}
& \klstd{9.254}{0.486}
& \klstd{8.331}{0.415}
& \klstd{8.177}{1.017}
& \klstd{6.399}{1.763}
& \klstd{5.545}{1.195}
& \klstd{4.801}{1.391}
& \klstd{\mathbf{4.346}}{1.101} \\

& Funnel
& \klstd{10.331}{0.073}
& \klstd{\mathbf{7.110}}{0.328}
& \klstd{10.671}{0.397}
& \klstd{13.264}{1.533}
& \klstd{13.766}{0.583}
& \klstd{14.424}{0.430}
& \klstd{14.322}{0.513}
& \klstd{14.617}{0.698}
& \klstd{15.645}{1.036}
& \klstd{20.068}{0.902} \\

& Ring
& \klstd{12.870}{0.052}
& \klstd{7.555}{0.074}
& \klstd{10.416}{0.190}
& \klstd{10.280}{0.426}
& \klstd{9.159}{0.646}
& \klstd{8.074}{0.895}
& \klstd{\mathbf{5.483}}{1.505}
& \klstd{5.603}{1.437}
& \klstd{5.543}{1.176}
& \klstd{21.898}{0.972} \\

\bottomrule
\end{tabular}%
}
\end{table*}

\begin{table*}[t]
\centering

\caption{
Complete ablation results for ancilla size $E$ on
forward KL divergence
(mean $\pm$ std over five runs; lower is better).
Bold indicates the lowest mean in each row.
A dash denotes an untested setting.
}

\label{tab:ablation_ancilla_full}
\scriptsize
\setlength{\tabcolsep}{4pt}
\resizebox{\textwidth}{!}{%
\begin{tabular}{cl*{11}{c}}
\toprule
\(D\) & Target
& \(E=0\) & \(E=1\) & \(E=2\) & \(E=3\) & \(E=4\)
& \(E=5\) & \(E=6\) & \(E=7\) & \(E=8\) & \(E=9\)
& \(E=10\) \\
\midrule

\multirow{4}{*}{5}
& X-shape
& \klstd{0.436}{0.162}
& \klstd{0.350}{0.050}
& \klstd{0.388}{0.040}
& \klstd{0.392}{0.052}
& \klstd{0.396}{0.038}
& \klstd{\mathbf{0.349}}{0.016}
& -- & -- & -- & -- & -- \\

& GMM3
& \klstd{\mathbf{0.173}}{0.023}
& \klstd{0.195}{0.021}
& \klstd{0.202}{0.033}
& \klstd{0.206}{0.031}
& \klstd{0.203}{0.014}
& \klstd{0.214}{0.076}
& -- & -- & -- & -- & -- \\

& Funnel
& \klstd{3.131}{0.553}
& \klstd{2.985}{0.562}
& \klstd{3.763}{0.782}
& \klstd{3.434}{0.811}
& \klstd{3.193}{0.342}
& \klstd{\mathbf{2.670}}{0.714}
& -- & -- & -- & -- & -- \\

& Ring
& \klstd{1.054}{0.143}
& \klstd{0.967}{0.066}
& \klstd{1.065}{0.204}
& \klstd{1.083}{0.180}
& \klstd{\mathbf{0.927}}{0.035}
& \klstd{1.046}{0.123}
& -- & -- & -- & -- & -- \\

\midrule

\multirow{4}{*}{10}
& X-shape
& \klstd{\mathbf{3.231}}{0.224}
& \klstd{3.382}{0.142}
& \klstd{3.253}{0.188}
& \klstd{3.334}{0.135}
& \klstd{3.339}{0.250}
& \klstd{3.415}{0.195}
& \klstd{3.277}{0.241}
& \klstd{3.321}{0.100}
& \klstd{3.385}{0.177}
& \klstd{3.470}{0.198}
& \klstd{3.366}{0.248} \\

& GMM3
& \klstd{3.405}{0.343}
& \klstd{3.283}{0.269}
& \klstd{3.216}{0.099}
& \klstd{3.309}{0.104}
& \klstd{3.250}{0.125}
& \klstd{3.206}{0.255}
& \klstd{3.340}{0.324}
& \klstd{3.239}{0.184}
& \klstd{\mathbf{3.091}}{0.166}
& \klstd{3.244}{0.206}
& \klstd{3.234}{0.136} \\

& Funnel
& \klstd{9.052}{1.045}
& \klstd{8.781}{0.682}
& \klstd{8.772}{0.581}
& \klstd{9.151}{0.397}
& \klstd{8.603}{0.808}
& \klstd{8.688}{1.054}
& \klstd{9.063}{0.322}
& \klstd{8.614}{0.840}
& \klstd{9.084}{0.486}
& \klstd{8.680}{0.808}
& \klstd{\mathbf{7.870}}{0.844} \\

& Ring
& \klstd{3.948}{0.187}
& \klstd{4.086}{0.398}
& \klstd{3.956}{0.308}
& \klstd{3.912}{0.288}
& \klstd{\mathbf{3.829}}{0.191}
& \klstd{3.958}{0.117}
& \klstd{3.966}{0.099}
& \klstd{4.073}{0.254}
& \klstd{3.953}{0.249}
& \klstd{4.063}{0.122}
& \klstd{4.052}{0.126} \\

\midrule

\multirow{4}{*}{20}
& X-shape
& \klstd{8.294}{1.607}
& \klstd{7.664}{1.668}
& \klstd{7.947}{1.502}
& \klstd{8.764}{1.985}
& \klstd{8.093}{0.189}
& \klstd{\mathbf{6.971}}{1.496}
& \klstd{8.519}{0.607}
& \klstd{8.062}{1.324}
& \klstd{7.306}{1.071}
& \klstd{8.374}{1.411}
& \klstd{8.491}{1.104} \\

& GMM3
& \klstd{8.021}{0.862}
& \klstd{7.944}{1.242}
& \klstd{7.916}{1.518}
& \klstd{8.068}{0.772}
& \klstd{8.552}{0.452}
& \klstd{8.500}{1.484}
& \klstd{8.646}{0.715}
& \klstd{8.308}{0.442}
& \klstd{\mathbf{7.648}}{1.596}
& \klstd{8.213}{0.556}
& \klstd{8.154}{1.810} \\

& Funnel
& \klstd{13.143}{1.655}
& \klstd{13.480}{0.675}
& \klstd{12.637}{1.569}
& \klstd{13.677}{0.840}
& \klstd{13.661}{0.550}
& \klstd{12.756}{1.658}
& \klstd{13.336}{1.518}
& \klstd{13.543}{0.657}
& \klstd{\mathbf{12.345}}{1.309}
& \klstd{13.405}{1.369}
& \klstd{13.843}{1.085} \\

& Ring
& \klstd{8.080}{1.967}
& \klstd{7.945}{1.077}
& \klstd{8.285}{1.318}
& \klstd{8.726}{1.454}
& \klstd{7.829}{1.846}
& \klstd{8.691}{1.005}
& \klstd{7.532}{1.904}
& \klstd{9.348}{0.442}
& \klstd{8.062}{1.856}
& \klstd{9.374}{0.495}
& \klstd{\mathbf{7.407}}{1.978} \\

\bottomrule
\end{tabular}%
}
\end{table*}
















\clearpage
\section*{NeurIPS Paper Checklist}

\begin{enumerate}

\item {\bf Claims}
    \item[] Question: Do the main claims made in the abstract and introduction accurately reflect the paper's contributions and scope?
    \item[] Answer: \answerYes{} 
    \item[] Justification: The abstract and introduction accurately summarize the
paper's scope: a density-operator formulation for degenerate low-energy
eigenspaces, an MPO-motivated QTN parameterization for tractable computation,
and empirical evaluation on synthetic targets, Bayesian posterior benchmarks,
and ablations. The claims are framed as methodological and empirical, with
scalability tied to local tensor-network structure rather than a general
worst-case guarantee.
    \item[] Guidelines:
    \begin{itemize}
        \item The answer \answerNA{} means that the abstract and introduction do not include the claims made in the paper.
        \item The abstract and/or introduction should clearly state the claims made, including the contributions made in the paper and important assumptions and limitations. A \answerNo{} or \answerNA{} answer to this question will not be perceived well by the reviewers. 
        \item The claims made should match theoretical and experimental results, and reflect how much the results can be expected to generalize to other settings. 
        \item It is fine to include aspirational goals as motivation as long as it is clear that these goals are not attained by the paper. 
    \end{itemize}

\item {\bf Limitations}
    \item[] Question: Does the paper discuss the limitations of the work performed by the authors?
    \item[] Answer: \answerYes{} 
    \item[] Justification: The paper discusses the main limitation that QuanVI relies
on local structure for efficient tensor-network contractions. It also notes that
larger local windows can improve expressiveness but increase memory and
runtime, as shown in the ablation studies. The conclusion further identifies
extensions to richer dependency graphs and adaptive tensor-network architectures
as future work.
    \item[] Guidelines:
    \begin{itemize}
        \item The answer \answerNA{} means that the paper has no limitation while the answer \answerNo{} means that the paper has limitations, but those are not discussed in the paper. 
        \item The authors are encouraged to create a separate ``Limitations'' section in their paper.
        \item The paper should point out any strong assumptions and how robust the results are to violations of these assumptions (e.g., independence assumptions, noiseless settings, model well-specification, asymptotic approximations only holding locally). The authors should reflect on how these assumptions might be violated in practice and what the implications would be.
        \item The authors should reflect on the scope of the claims made, e.g., if the approach was only tested on a few datasets or with a few runs. In general, empirical results often depend on implicit assumptions, which should be articulated.
        \item The authors should reflect on the factors that influence the performance of the approach. For example, a facial recognition algorithm may perform poorly when image resolution is low or images are taken in low lighting. Or a speech-to-text system might not be used reliably to provide closed captions for online lectures because it fails to handle technical jargon.
        \item The authors should discuss the computational efficiency of the proposed algorithms and how they scale with dataset size.
        \item If applicable, the authors should discuss possible limitations of their approach to address problems of privacy and fairness.
        \item While the authors might fear that complete honesty about limitations might be used by reviewers as grounds for rejection, a worse outcome might be that reviewers discover limitations that aren't acknowledged in the paper. The authors should use their best judgment and recognize that individual actions in favor of transparency play an important role in developing norms that preserve the integrity of the community. Reviewers will be specifically instructed to not penalize honesty concerning limitations.
    \end{itemize}

\item {\bf Theory assumptions and proofs}
    \item[] Question: For each theoretical result, does the paper provide the full set of assumptions and a complete (and correct) proof?
    \item[] Answer: \answerNA{} 
    \item[] Justification: The paper does not present formal theorem-style theoretical results. It provides derivations of the density-operator formulation and complexity analysis under the stated locality assumption, which are used to motivate and analyze the method.
    \item[] Guidelines:
    \begin{itemize}
        \item The answer \answerNA{} means that the paper does not include theoretical results. 
        \item All the theorems, formulas, and proofs in the paper should be numbered and cross-referenced.
        \item All assumptions should be clearly stated or referenced in the statement of any theorems.
        \item The proofs can either appear in the main paper or the supplemental material, but if they appear in the supplemental material, the authors are encouraged to provide a short proof sketch to provide intuition. 
        \item Inversely, any informal proof provided in the core of the paper should be complemented by formal proofs provided in appendix or supplemental material.
        \item Theorems and Lemmas that the proof relies upon should be properly referenced. 
    \end{itemize}

    \item {\bf Experimental result reproducibility}
    \item[] Question: Does the paper fully disclose all the information needed to reproduce the main experimental results of the paper to the extent that it affects the main claims and/or conclusions of the paper (regardless of whether the code and data are provided or not)?
    \item[] Answer: \answerYes{} 
    \item[] Justification: The paper provides the information needed to reproduce the
main experiments, including target construction, PosteriorDB benchmark mapping,
evaluation metrics, baselines, hyperparameters, number of runs, ablation
settings, and computational resources.
    \item[] Guidelines:
    \begin{itemize}
        \item The answer \answerNA{} means that the paper does not include experiments.
        \item If the paper includes experiments, a \answerNo{} answer to this question will not be perceived well by the reviewers: Making the paper reproducible is important, regardless of whether the code and data are provided or not.
        \item If the contribution is a dataset and\slash or model, the authors should describe the steps taken to make their results reproducible or verifiable. 
        \item Depending on the contribution, reproducibility can be accomplished in various ways. For example, if the contribution is a novel architecture, describing the architecture fully might suffice, or if the contribution is a specific model and empirical evaluation, it may be necessary to either make it possible for others to replicate the model with the same dataset, or provide access to the model. In general. releasing code and data is often one good way to accomplish this, but reproducibility can also be provided via detailed instructions for how to replicate the results, access to a hosted model (e.g., in the case of a large language model), releasing of a model checkpoint, or other means that are appropriate to the research performed.
        \item While NeurIPS does not require releasing code, the conference does require all submissions to provide some reasonable avenue for reproducibility, which may depend on the nature of the contribution. For example
        \begin{enumerate}
            \item If the contribution is primarily a new algorithm, the paper should make it clear how to reproduce that algorithm.
            \item If the contribution is primarily a new model architecture, the paper should describe the architecture clearly and fully.
            \item If the contribution is a new model (e.g., a large language model), then there should either be a way to access this model for reproducing the results or a way to reproduce the model (e.g., with an open-source dataset or instructions for how to construct the dataset).
            \item We recognize that reproducibility may be tricky in some cases, in which case authors are welcome to describe the particular way they provide for reproducibility. In the case of closed-source models, it may be that access to the model is limited in some way (e.g., to registered users), but it should be possible for other researchers to have some path to reproducing or verifying the results.
        \end{enumerate}
    \end{itemize}

\item {\bf Open access to data and code}
    \item[] Question: Does the paper provide open access to the data and code, with sufficient instructions to faithfully reproduce the main experimental results, as described in supplemental material?
    \item[] Answer: \answerYes{} 
    \item[] Justification: The code needed to reproduce the main experimental results
is included in the supplementary material.
    \item[] Guidelines:
    \begin{itemize}
        \item The answer \answerNA{} means that paper does not include experiments requiring code.
        \item Please see the NeurIPS code and data submission guidelines (\url{https://neurips.cc/public/guides/CodeSubmissionPolicy}) for more details.
        \item While we encourage the release of code and data, we understand that this might not be possible, so \answerNo{} is an acceptable answer. Papers cannot be rejected simply for not including code, unless this is central to the contribution (e.g., for a new open-source benchmark).
        \item The instructions should contain the exact command and environment needed to run to reproduce the results. See the NeurIPS code and data submission guidelines (\url{https://neurips.cc/public/guides/CodeSubmissionPolicy}) for more details.
        \item The authors should provide instructions on data access and preparation, including how to access the raw data, preprocessed data, intermediate data, and generated data, etc.
        \item The authors should provide scripts to reproduce all experimental results for the new proposed method and baselines. If only a subset of experiments are reproducible, they should state which ones are omitted from the script and why.
        \item At submission time, to preserve anonymity, the authors should release anonymized versions (if applicable).
        \item Providing as much information as possible in supplemental material (appended to the paper) is recommended, but including URLs to data and code is permitted.
    \end{itemize}

\item {\bf Experimental setting/details}
    \item[] Question: Does the paper specify all the training and test details (e.g., data splits, hyperparameters, how they were chosen, type of optimizer) necessary to understand the results?
    \item[] Answer: \answerYes{} 
    \item[] Justification: The paper reports the relevant training and evaluation
details, including target construction, evaluation metrics, baselines,
hyperparameters, optimization budgets, number of runs, and computational
resources, with additional settings provided in the appendix.
    \item[] Guidelines:
    \begin{itemize}
        \item The answer \answerNA{} means that the paper does not include experiments.
        \item The experimental setting should be presented in the core of the paper to a level of detail that is necessary to appreciate the results and make sense of them.
        \item The full details can be provided either with the code, in appendix, or as supplemental material.
    \end{itemize}

\item {\bf Experiment statistical significance}
    \item[] Question: Does the paper report error bars suitably and correctly defined or other appropriate information about the statistical significance of the experiments?
    \item[] Answer: \answerYes{} 
    \item[] Justification: The paper reports means and standard deviations over
repeated runs for the main quantitative comparisons and ablation studies where
multiple runs are performed. These statistics provide uncertainty information
for the empirical comparisons, although some large-scale settings use a limited
number of repetitions due to computational cost.
    \item[] Guidelines:
    \begin{itemize}
        \item The answer \answerNA{} means that the paper does not include experiments.
        \item The authors should answer \answerYes{} if the results are accompanied by error bars, confidence intervals, or statistical significance tests, at least for the experiments that support the main claims of the paper.
        \item The factors of variability that the error bars are capturing should be clearly stated (for example, train/test split, initialization, random drawing of some parameter, or overall run with given experimental conditions).
        \item The method for calculating the error bars should be explained (closed form formula, call to a library function, bootstrap, etc.)
        \item The assumptions made should be given (e.g., Normally distributed errors).
        \item It should be clear whether the error bar is the standard deviation or the standard error of the mean.
        \item It is OK to report 1-sigma error bars, but one should state it. The authors should preferably report a 2-sigma error bar than state that they have a 96\% CI, if the hypothesis of Normality of errors is not verified.
        \item For asymmetric distributions, the authors should be careful not to show in tables or figures symmetric error bars that would yield results that are out of range (e.g., negative error rates).
        \item If error bars are reported in tables or plots, the authors should explain in the text how they were calculated and reference the corresponding figures or tables in the text.
    \end{itemize}

\item {\bf Experiments compute resources}
    \item[] Question: For each experiment, does the paper provide sufficient information on the computer resources (type of compute workers, memory, time of execution) needed to reproduce the experiments?
    \item[] Answer: \answerYes{} 
    \item[] Justification: The paper provides the main compute-resource information
needed to contextualize the experiments, including CPU types, memory, and which
resources were used for the synthetic, posterior, and ablation experiments.
Detailed runtime comparisons are reported for settings where computational cost
is directly analyzed.
    \item[] Guidelines:
    \begin{itemize}
        \item The answer \answerNA{} means that the paper does not include experiments.
        \item The paper should indicate the type of compute workers CPU or GPU, internal cluster, or cloud provider, including relevant memory and storage.
        \item The paper should provide the amount of compute required for each of the individual experimental runs as well as estimate the total compute. 
        \item The paper should disclose whether the full research project required more compute than the experiments reported in the paper (e.g., preliminary or failed experiments that didn't make it into the paper). 
    \end{itemize}
    
\item {\bf Code of ethics}
    \item[] Question: Does the research conducted in the paper conform, in every respect, with the NeurIPS Code of Ethics \url{https://neurips.cc/public/EthicsGuidelines}?
    \item[] Answer: \answerYes{} 
    \item[] Justification: The authors have reviewed the NeurIPS Code of Ethics and
confirm that the research complies with the ethical standards and submission
guidelines of the conference.
    \item[] Guidelines:
    \begin{itemize}
        \item The answer \answerNA{} means that the authors have not reviewed the NeurIPS Code of Ethics.
        \item If the authors answer \answerNo, they should explain the special circumstances that require a deviation from the Code of Ethics.
        \item The authors should make sure to preserve anonymity (e.g., if there is a special consideration due to laws or regulations in their jurisdiction).
    \end{itemize}

\item {\bf Broader impacts}
    \item[] Question: Does the paper discuss both potential positive societal impacts and negative societal impacts of the work performed?
    \item[] Answer: \answerNA{} 
    \item[] Justification: The work is methodological and does not present a broader
impacts discussion beyond the standard scope of algorithmic development and
benchmark evaluation.
    \item[] Guidelines:
    \begin{itemize}
        \item The answer \answerNA{} means that there is no societal impact of the work performed.
        \item If the authors answer \answerNA{} or \answerNo, they should explain why their work has no societal impact or why the paper does not address societal impact.
        \item Examples of negative societal impacts include potential malicious or unintended uses (e.g., disinformation, generating fake profiles, surveillance), fairness considerations (e.g., deployment of technologies that could make decisions that unfairly impact specific groups), privacy considerations, and security considerations.
        \item The conference expects that many papers will be foundational research and not tied to particular applications, let alone deployments. However, if there is a direct path to any negative applications, the authors should point it out. For example, it is legitimate to point out that an improvement in the quality of generative models could be used to generate Deepfakes for disinformation. On the other hand, it is not needed to point out that a generic algorithm for optimizing neural networks could enable people to train models that generate Deepfakes faster.
        \item The authors should consider possible harms that could arise when the technology is being used as intended and functioning correctly, harms that could arise when the technology is being used as intended but gives incorrect results, and harms following from (intentional or unintentional) misuse of the technology.
        \item If there are negative societal impacts, the authors could also discuss possible mitigation strategies (e.g., gated release of models, providing defenses in addition to attacks, mechanisms for monitoring misuse, mechanisms to monitor how a system learns from feedback over time, improving the efficiency and accessibility of ML).
    \end{itemize}
    
\item {\bf Safeguards}
    \item[] Question: Does the paper describe safeguards that have been put in place for responsible release of data or models that have a high risk for misuse (e.g., pre-trained language models, image generators, or scraped datasets)?
    \item[] Answer: \answerNA{} 
    \item[] Justification: The paper does not release data or models with high
misuse risk, such as pretrained generative models, language models, image
generators, or scraped datasets. The released code supports methodological
experiments on synthetic targets and public benchmark models.
    \item[] Guidelines:
    \begin{itemize}
        \item The answer \answerNA{} means that the paper poses no such risks.
        \item Released models that have a high risk for misuse or dual-use should be released with necessary safeguards to allow for controlled use of the model, for example by requiring that users adhere to usage guidelines or restrictions to access the model or implementing safety filters. 
        \item Datasets that have been scraped from the Internet could pose safety risks. The authors should describe how they avoided releasing unsafe images.
        \item We recognize that providing effective safeguards is challenging, and many papers do not require this, but we encourage authors to take this into account and make a best faith effort.
    \end{itemize}

\item {\bf Licenses for existing assets}
    \item[] Question: Are the creators or original owners of assets (e.g., code, data, models), used in the paper, properly credited and are the license and terms of use explicitly mentioned and properly respected?
    \item[] Answer: \answerYes{} 
    \item[] Justification: The paper credits the existing assets and prior work used
in the experiments, including \textsc{PosteriorDB}, EigenVI, and the baseline
methods. Public benchmark assets are used according to their intended research
purpose, and the corresponding sources are cited in the main text and appendix.
    \item[] Guidelines:
    \begin{itemize}
        \item The answer \answerNA{} means that the paper does not use existing assets.
        \item The authors should cite the original paper that produced the code package or dataset.
        \item The authors should state which version of the asset is used and, if possible, include a URL.
        \item The name of the license (e.g., CC-BY 4.0) should be included for each asset.
        \item For scraped data from a particular source (e.g., website), the copyright and terms of service of that source should be provided.
        \item If assets are released, the license, copyright information, and terms of use in the package should be provided. For popular datasets, \url{paperswithcode.com/datasets} has curated licenses for some datasets. Their licensing guide can help determine the license of a dataset.
        \item For existing datasets that are re-packaged, both the original license and the license of the derived asset (if it has changed) should be provided.
        \item If this information is not available online, the authors are encouraged to reach out to the asset's creators.
    \end{itemize}

\item {\bf New assets}
    \item[] Question: Are new assets introduced in the paper well documented and is the documentation provided alongside the assets?
    \item[] Answer: \answerNA{} 
    \item[] Justification: The paper does not introduce new datasets, models, or benchmark assets. The supplementary code is provided only to reproduce the reported experiments.
    \item[] Guidelines:
    \begin{itemize}
        \item The answer \answerNA{} means that the paper does not release new assets.
        \item Researchers should communicate the details of the dataset\slash code\slash model as part of their submissions via structured templates. This includes details about training, license, limitations, etc. 
        \item The paper should discuss whether and how consent was obtained from people whose asset is used.
        \item At submission time, remember to anonymize your assets (if applicable). You can either create an anonymized URL or include an anonymized zip file.
    \end{itemize}

\item {\bf Crowdsourcing and research with human subjects}
    \item[] Question: For crowdsourcing experiments and research with human subjects, does the paper include the full text of instructions given to participants and screenshots, if applicable, as well as details about compensation (if any)? 
    \item[] Answer: \answerNA{} 
    \item[] Justification:The paper does not involve crowdsourcing experiments or
research with human subjects.
    \item[] Guidelines:
    \begin{itemize}
        \item The answer \answerNA{} means that the paper does not involve crowdsourcing nor research with human subjects.
        \item Including this information in the supplemental material is fine, but if the main contribution of the paper involves human subjects, then as much detail as possible should be included in the main paper. 
        \item According to the NeurIPS Code of Ethics, workers involved in data collection, curation, or other labor should be paid at least the minimum wage in the country of the data collector. 
    \end{itemize}

\item {\bf Institutional review board (IRB) approvals or equivalent for research with human subjects}
    \item[] Question: Does the paper describe potential risks incurred by study participants, whether such risks were disclosed to the subjects, and whether Institutional Review Board (IRB) approvals (or an equivalent approval/review based on the requirements of your country or institution) were obtained?
    \item[] Answer: \answerNA{} 
    \item[] Justification: The paper does not involve research with human subjects,
so IRB approval or equivalent review is not applicable.
    \item[] Guidelines:
    \begin{itemize}
        \item The answer \answerNA{} means that the paper does not involve crowdsourcing nor research with human subjects.
        \item Depending on the country in which research is conducted, IRB approval (or equivalent) may be required for any human subjects research. If you obtained IRB approval, you should clearly state this in the paper. 
        \item We recognize that the procedures for this may vary significantly between institutions and locations, and we expect authors to adhere to the NeurIPS Code of Ethics and the guidelines for their institution. 
        \item For initial submissions, do not include any information that would break anonymity (if applicable), such as the institution conducting the review.
    \end{itemize}

\item {\bf Declaration of LLM usage}
    \item[] Question: Does the paper describe the usage of LLMs if it is an important, original, or non-standard component of the core methods in this research? Note that if the LLM is used only for writing, editing, or formatting purposes and does \emph{not} impact the core methodology, scientific rigor, or originality of the research, declaration is not required.
    \item[] Answer: \answerNA{} 
    \item[] Justification: LLMs were not used as an important, original, or
non-standard component of the core methodology, experiments, or scientific
claims. Any use of LLMs, if applicable, was limited to writing, editing, or
formatting assistance.
    \item[] Guidelines:
    \begin{itemize}
        \item The answer \answerNA{} means that the core method development in this research does not involve LLMs as any important, original, or non-standard components.
        \item Please refer to our LLM policy in the NeurIPS handbook for what should or should not be described.
    \end{itemize}

\end{enumerate}

\end{document}